\documentclass{article}
     \usepackage[preprint,nonatbib]{neurips_2024}

\usepackage[utf8]{inputenc} % allow utf-8 input
\usepackage[T1]{fontenc}    % use 8-bit T1 fonts
\usepackage{hyperref}       % hyperlinks
\usepackage{url}            % simple URL typesetting
\usepackage{booktabs}       % professional-quality tables
\usepackage{amsfonts}       % blackboard math symbols
\usepackage{nicefrac}       % compact symbols for 1/2, etc.
\usepackage{microtype}      % microtypography
\usepackage{xcolor}         % colors

\usepackage{graphicx}
\usepackage{amsmath,amsthm,amssymb}
\usepackage[caption=false,farskip=0pt,labelfont={bf}]{subfig} 
\usepackage{multirow}
\usepackage{enumitem}
\usepackage{makecell}
\usepackage{arydshln}

\usepackage{algorithm}

\usepackage{algpseudocode}

\usepackage{amsmath,amsthm,amssymb}
\usepackage{color}

\newtheorem{theorem}{Theorem}%[section]
\newtheorem{definition}{Definition}%[section]

\newcommand{\beq}{\begin{equation}}
\newcommand{\eeq}{\end{equation}}
\newcommand{\beqs}{\begin{eqnarray}}
\newcommand{\eeqs}{\end{eqnarray}}
\newcommand{\barr}{\begin{array}}
	\newcommand{\earr}{\end{array}}
\newcommand{\bali}{\begin{aligned}}
	\newcommand{\eali}{\end{aligned}}

\newcommand{\Lc}[0]{\ensuremath{\mathcal{L}} }

\newcommand{\Nc}[0]{\ensuremath{\mathcal{N}} }

\newcommand{\Ebb}[0]{\ensuremath{\mathbb{E}} }

\newcommand{\Lbb}[0]{\ensuremath{\mathbb{L}} }

\newcommand{\Rbb}[0]{\ensuremath{\mathbb{R}} }
\newcommand{\Sbb}[0]{\ensuremath{\mathbb{S}} }
\newcommand{\Tbb}[0]{\ensuremath{\mathbb{T}} }

\newcommand{\Zbb}[0]{\ensuremath{\mathbb{Z}} }

\newcommand{\ie}[0]{\emph{i.e., }}

\newcommand{\eg}[0]{\emph{e.g., }}

\newcommand{\etc}[0]{\emph{etc. }}

\newcommand{\wrt}[0]{\emph{w.r.t. }}
\newcommand{\iid}[0]{\emph{i.i.d. }}

\newcommand{\Amat}[0]{\ensuremath{{\bf A}} }
\newcommand{\Bmat}[0]{\ensuremath{{\bf B}} }

\newcommand{\Imat}[0]{\ensuremath{{\bf I}} }

\newcommand{\Lmat}[0]{\ensuremath{{\bf L}} }

\newcommand{\Xmat}[0]{\ensuremath{{\bf X}} }

\newcommand{\nv}[0]{\ensuremath{\boldsymbol{n}} }

\newcommand{\xv}[0]{\ensuremath{\boldsymbol{x}} }
\newcommand{\yv}[0]{\ensuremath{\boldsymbol{y}} }
\newcommand{\zv}[0]{\ensuremath{\boldsymbol{z}} }

\newcommand{\Pv}[0]{\ensuremath{\boldsymbol{P}} }

\newcommand{\Xv}[0]{\ensuremath{\boldsymbol{X}} }

\newcommand{\Sigmamat}[0]{\ensuremath{\boldsymbol{\Sigma}} }

\newcommand{\Omegamat}[0]{\ensuremath{\boldsymbol{\Omega}}}

\newcommand{\thetav}[0]{\ensuremath{\boldsymbol{\theta}} }

\newcommand{\muv}[0]{\ensuremath{\boldsymbol{\mu}} }

\newcommand{\phiv}[0]{\ensuremath{\boldsymbol{\phi}} }

\newcommand{\KL}[0]{\ensuremath{\mathrm{KL}} }
\newcommand{\JS}[0]{\ensuremath{\mathrm{JS}} }

\theoremstyle{plain}
\theoremstyle{definition}
\theoremstyle{remark}
\newtheorem{remark}[theorem]{Remark}

\title{Big Cooperative Learning}

\author{%
	Yulai Cong \\ % $^{*}$
	Sun Yat-sen University \\
	\texttt{yulaicong@gmail.com} \\
}

\begin{document}

\maketitle

%\vspace{-5mm}
\begin{abstract}

Cooperation plays a pivotal role in the evolution of human intelligence; moreover, it also underlies the recent revolutionary advancement of artificial intelligence (AI) that is driven by foundation models. 
Specifically, we reveal that the training of foundation models can be interpreted as a form of big cooperative learning (\textit{abbr.} big learning), where massive learning individuals/tasks \emph{cooperate} to approach the unique essence of data from diverse perspectives of data prediction, leveraging a universal model. 
The presented big learning therefore unifies most training objectives of foundation models within a consistent framework, where their underlying assumptions are exposed simultaneously.
We design tailored simulations to demonstrate the principle of big learning, based on which we provide learning-perspective justifications for the successes of foundation models, with interesting side-products. 
Furthermore, we reveal that big learning is a new dimension for upgrading conventional machine learning paradigms, valuable for endowing reinvigorations to associated applications;
as an illustrative example, we propose the BigLearn-GAN, which is a novel adversarially-trained foundation model with versatile data sampling capabilities.
Code is available at \texttt{https://github.com/YulaiCong/BigCooperativeLearning}.

\end{abstract}

\section{Introduction}

Cooperation is essential to the survival of human society and to the formulation and evolution of human intelligence.
``Without the cooperation of its members society cannot survive, and the society of man has survived because the cooperativeness of its members made survival possible .... It was not an advantageous individual here and there who did so, but the group'' (Ashley Montagu, 1965) \cite{roger1994overview}.
Regarding the formulation of human intelligence, \emph{cooperative learning} is an effective and efficient educational tool extensively researched by anthropologists and educational psychologists \cite{slavin1980cooperative,johnson1987learning,johnson1989cooperation,johnson1991joining}, where \emph{different individuals} (often students with different capabilities) cooperate with each other to achieve \emph{a common goal}.

The world is experiencing a groundbreaking AI revolution with cooperation among scientists and the rise of big/foundation models \cite{bommasani2021opportunities,yuan2022roadmap}, such as the GPT series \cite{GPT4,ChatGPT,ouyang2022training,brown2020language}, Sora \cite{videoworldsimulators2024}, Geminis \cite{reid2024gemini,team2023gemini}, DALL-Es \cite{ramesh2021zero,ramesh2022hierarchical,DallE3}, and BERTs \cite{devlin2018bert,lan2019albert,liu2019roberta}, which brings advanced AI capabilities to diverse application domains with impressive robustness \cite{stickland2019bert}, adaptability \cite{he2021masked}, and generalization \cite{ramesh2021zero}, lighting up the way towards Artificial General Intelligence (AGI).

In addition to advancing human intelligence, we reveal that cooperation also plays a crucial role in the extraordinary successes of foundation models.
Specifically, the training of most foundation models (such as mask-and-predict \cite{devlin2018bert} and next-token-prediction \cite{brown2020language}) can be interpreted as \textbf{big cooperative learning} (abbreviated as \textbf{big learning}), where \textbf{massive learning individuals} (\eg learning tasks with various masks/preceding-contexts for mask-and-predict/next-token-prediction) \textbf{have different and diverse characteristics} (\ie focus on different conditional data predictions) \textbf{but share a common goal} (\ie to approach the unique essence of data with a universal set of model parameters).
We will show that the presented big cooperative learning contains most training objectives of foundation models as special cases and therefore it manifests as a unified and general learning framework for foundation models, which is considered crucial for their further improvement  \cite{tamkin2021dabs,bommasani2021opportunities,yuan2022roadmap}.

%demonstrates how a foundation model learns/summarizes the essential data information within its parameters. 
%and might be a milestone for the foundation model community

Given trustworthy data and a universal model with sufficient capacity and flexibility, big cooperative learning exhaustively exploits data information from diverse perspectives to form the aforementioned massive learning tasks, whose local optima unstably vary (with the task) but their global optima (\ie the common goal) are stably the same. 
We will bring to light that big cooperative learning leverages cooperation among those massive learning tasks to deliver a remarkable power of exploration that overlooks local optima and focuses on the global optima.

%%%%%%%%%%%%%%%%%%%%%%%%%%       Contribtution
%	We propose the general concept of Big Cooperative Learning
%	We reveal that big learning unify most training objectives of foundation models within one consistent framework. 
%		we reveal explicit the assumptions that each FM made in the context of big learning, assumption further improvement.
%	
%	Our other contributions are listed as follows.
%		We leverage tailored simulation to analyze demonstrate big learning lightweight, which explicitly and justify success of FM from objective perspective.
%		
%		We reveal that the big learning consistent train/inference goal as member task, min the gap, better learning, and directly form capabilities.
%			can be leveraged to deliver \emph{many/all} joint, conditional, and marginal data sampling capabilities with one universal foundation model.
%			Those capabilities, in general settings, can manifest as classification, generation, completion/in-painting, \etc
%		
%		We leverage the big learning principle to upgrade the conventional generative adversarial net (GAN) into its big-learning variant termed the BigLearn-GAN, which is a novel adversarially-trained foundation model. 
%		
%		We empirically demonstrate that big learning ($i$) is feasible, ($ii$) delivers good model generalization, and ($iii$) can serve as a better strategy for finetuning foundation models.
%	

Our contributions are listed as follows.
\begin{itemize}[leftmargin=5mm]%[leftmargin=*,topsep=0.cm,itemsep=0.cm,partopsep=0cm,parsep=0cm]
    
    \item We present the general concept of big cooperative learning (\textit{abbr.} big learning) to unify most of the training objectives of foundation models within a unified learning framework; based on this, we expose and analyze the underlying assumptions made by existing foundation models.
    
    \item We design tailored simulations to demonstrate the principle of big learning in a lightweight way, based on which we explicitly justify the successes of foundation models from the perspective of learning and provide interesting side-products that potentially benefit their improvement.
    
    \item We bring to light that big learning is a new dimension for upgrading conventional machine learning paradigms, demonstrating knowledge feedback from cutting-edge foundation models to conventional machine learning and endowing reinvigorations to numerous associated applications.
    
	\item As an illustrative example, we leverage the presented big learning to upgrade the conventional generative adversarial net (GAN) into its big-learning variant termed BigLearn-GAN, which manifests as a novel adversarially-trained foundation model. 
    
    \item We empirically show that the massive learning tasks of big learning naturally deliver the associated valuable data-sampling capabilities during testing; these capabilities, in a general multi-modal setting, can manifest as versatile cross-modal generations.
    
%    (\ie joint, marginal, and conditional data sampling in transformed domains), in a general multi-modal setting, can manifest as versatile cross-modal generations.
    
\end{itemize}

\section{Preliminary}

\textbf{Big/Foundation models.} 
Taking shape in the field of natural language processing (NLP), foundation models have dramatically changed AI-related research and applications, sparking a groundbreaking AI revolution that is sweeping the world \cite{bommasani2021opportunities,yuan2022roadmap}.
Representative foundation models include the BERT series \cite{devlin2018bert,lan2019albert,stickland2019bert,liu2019roberta,joshi2020spanbert}, which reform NLP applications, the GPT series \cite{radford2019language,brown2020language,ouyang2022training,ChatGPT,GPT4}, which give rise to a worldwide explosion of large language models \cite{team2023gemini,reid2024gemini,claude3,llama3modelcard,abdin2024phi}, and multi-modal foundation models \cite{rombach2022high,bao2023one,DallE3,videoworldsimulators2024}, which initiate a new revolution of AI-Generated Content (AIGC).

Foundation models are mainly trained/pretrained under the maximum-likelihood-learning principle with either mask-and-predict (\ie masked auto-encoding or masked language modeling) \cite{devlin2018bert,he2021masked} or next-token-prediction (\ie auto-regressive/causal language modeling) \cite{radford2019language,brown2020language}.

Specifically, the mask-and-predict training seeks to optimize the universal set of parameters $\thetav$ via
\beq\label{eq:mask-and-predict}
	\max_{\thetav} \Ebb_{q(\xv) q(\Sbb)} \log p^{\text{MAE}}_{\thetav}(\xv_{\Sbb^{\complement}}|\xv_{\Sbb})
	\quad \text{or} \quad
	\max_{\thetav} \Ebb_{q(\xv_{\Sbb})} \Ebb_{q(\xv_{\Sbb^{\complement}}|\xv_{\Sbb})} \log p^{\text{MAE}}_{\thetav}(\xv_{\Sbb^{\complement}}|\xv_{\Sbb}),
\eeq
where $\xv \in \Rbb^{L\times D}$, sampled from the underlying data distribution $q(\xv)$,
%\footnote{
%	$q(\cdot)$ denotes a fixed PDF (like the data PDF $q(\xv)$ or the predefined $q(\Sbb,\Tbb)$ for sampling $(\Sbb,\Tbb)$), while $p_{\thetav}(\cdot)$ denotes a trainable model PDF. 
%}, 
denotes a sequence of $L$ tokens of dimension $D$ (\eg $\xv \in \Zbb^{6\times 1}$ may represent the $6$-word sentence of ``big learning is a general concept''), $\Sbb$ (sampled from a predefined $q(\Sbb)$) is a random subset of the token index set $\Lbb=\{1,\cdots,L\}$ (\eg $\Sbb=\{1,3,4,5\}$, its complement $\Sbb^{\complement}=\{2,6\}$, $\xv_{\Sbb}$ is ``big \_ is a general \_'', and $\xv_{\Sbb^{\complement}}$ is ``\_ learning \_ \_ \_ concept''), and the model $p^{\text{MAE}}_{\thetav}(\xv_{\Sbb^{\complement}}|\xv_{\Sbb})$ adopts the conditional independence assumption that ignores the dependency among the tokens of $\xv_{\Sbb^{\complement}}$, \ie
\beq
	p^{\text{MAE}}_{\thetav}(\xv_{\Sbb^{\complement}}|\xv_{\Sbb}) = \prod\nolimits_{t\in\Sbb^{\complement}} p^{\text{MAE}}_{\thetav}(\xv_{t}|\xv_{\Sbb}),
\eeq
where $p^{\text{MAE}}_{\thetav}(\xv_{t}|\xv_{\Sbb})$ is often modeled as a Categorical distribution, \ie $\text{Categorical}(x_{t}|\Pv_{\thetav}(\xv_{\Sbb}))$ with probabilities $\Pv_{\thetav}(\xv_{\Sbb})$, for discrete (text) token $x_{t}\in\Zbb^1$ \cite{devlin2018bert}, and a Gaussian distribution, \ie $\Nc(\xv_{t}|\muv_{\thetav}(\xv_{\Sbb}), \sigma^2)$ with mean $\muv_{\thetav}(\xv_{\Sbb})$ and variance $\sigma^2$, for continuous (image) token $\xv_{t}\in\Rbb^{D}$ \cite{he2021masked}.

Different from mask-and-predict, the next-token-prediction training optimizes the universal parameter set $\thetav$ in an auto-regressive manner via 
\beq\label{eq:next-token-prediction}
	\max_{\thetav} \Ebb_{q(\xv)} \sum\nolimits_{t=1}^L \log p^{\text{AR}}_{\thetav}(x_{t}|\xv_{{<t}})
	\quad \text{or} \quad
	\max_{\thetav} \Ebb_{q(\xv)q(t)} \log p^{\text{AR}}_{\thetav}(x_{t}|\xv_{{<t}}),
\eeq
where ${<t} \triangleq \{1,\cdots,t-1\}$, $\xv_{{<t}}$ denotes the tokens prior to the $t$-th token $x_{t}$, $q(t)=U[1,L]$ is a uniform distribution, and the universal parameter set $\thetav$ of the model $p^{\text{AR}}_{\thetav}(x_{t}|\xv_{{<t}})$ is shared across all next-token-prediction tasks with different $t$s.

Beyond the vanilla mask-and-predict and next-token-prediction, many generalized variants of them have been proposed for training/pretraining foundation models. 
For example, the permutation language modeling \cite{yang2019xlnet} leverages permutations to combine their advantages while overcoming their limitations from either the conditional independence assumption or the fixed forward factorization; 
MAE \cite{he2021masked} and MaskFeat \cite{wei2021masked} show that it's better to conduct the mask-and-predict training in normalized and HOG-transformed domains, respectively;
and VAR \cite{tian2024visual} similarly shows that replacing next-token-prediction with the transformed auto-regressive next-resolution-prediction results in SOTA image generation performance surpassing diffusion transformers.
We will show that all these training methods for foundation models share the same underlying learning principle that is summarized in the presented big cooperative learning.

\textbf{Transformers.} 
A sufficiently flexible network architecture capable of \emph{simultaneously} modeling versatile inputs and outputs (see Eqs. \eqref{eq:mask-and-predict} and \eqref{eq:next-token-prediction}) is essential for a successful foundation model.
Built on top of the flexible attention mechanism \cite{vaswani2017attention}, Transformers and their variants effectively address this challenge and have therefore served as the de facto architecture for foundation models.
Although a better architecture often leads to better performance, we will not focus on this in this paper, but rather on the learning theory of foundation models.

\textbf{Maximum-likelihood and adversarial learning} are two popular learning paradigms that train a model $p_{\thetav}(\xv)$ based on a collection of data samples from the underlying data distribution $q(\xv)$.
Specifically, maximum-likelihood learning is identical to minimizing the forward Kullback-Leibler (KL) divergence between $q(\xv)$ and $p_{\thetav}(\xv)$ because
\beq\label{eq:mle}
	\Ebb_{q(\xv)} [- \log p_{\thetav}(\xv)] = \KL[q(\xv)||p_{\thetav}(\xv)] - C,
\eeq
where $C=\Ebb_{q(\xv)} [\log q(\xv)]$ is a constant.
In comparison, adversarial learning (take the standard GAN \cite{goodfellow2014generative} as an example) minimizes the Jensen-Shannon (JS) divergence $\JS[q(\xv)||p_{\thetav}(\xv)]$, by training a generator $G_{\thetav}(\cdot)$ and a discriminator $\sigma[f_{\phiv}(\cdot)]$ in an adversarial manner via
\beq\label{eq:standard_GAN_loss}
	\min_{\thetav} \max_{\phiv} \Ebb_{q(\xv)} \log \sigma[f_{\phiv}(\xv)] + \Ebb_{p_{\thetav}(\xv)} \log (1-\sigma[f_{\phiv}(\xv)]) ,
\eeq
where $\sigma[\cdot]$ is the sigmoid function, $p_{\thetav}(\xv)$ is the model distribution with the generative process $\xv=G_{\thetav}(\zv), \zv\sim p(\zv)$, and $p(\zv)$ is an easy-to-sample (often normal) distribution.

%assumption:  p=q   with  application dependent
%
%fact: data have many perspective data capabilities but with the unique data essence. Many FM use part of that ... pretraining test gap
%	
%	existing FM pretraining exploit some potion of data perspective (to form the corresponding learning Individuals)
%	
%	MAE vs AR pretraining test gap  unform and consistence manner
%
%propose BL   train test
%	train test gap
%	
%	Special case summarize
%	
%Toy simulation to demonstrate BL cooperatiion among tasks learning individuals 
%
%BLGAN 
%
%Discuss :  data correctness trustworthy.    current state of data cleaning
%
%utilizes versatile sampling demonstrations to form learning individuals 
%
%a sample demonstrate versatile data sampling capabilities

\section{Big Cooperative Learning}
\label{sec:our_method}

\textbf{Assumptions.}
Throughout the paper, we assume trustworthy data samples from $q(\xv)$, sufficient model capacity of $p_{\thetav}(\xv)$, and the existence of an optimal $\thetav^{*}$ satisfying $p_{\thetav^{*}}(\xv)=q(\xv)$.
Note $p_{\thetav^{*}}(\xv)=q(\xv)$ means that the \emph{data essence} is implicitly summarized in $\thetav^{*}$ and that 
\begin{itemize}[leftmargin=5mm]	
	\item both sides (\eg \emph{joint} distributions of $\xv=[x_1,\cdots,x_L]^T \in \Rbb^{L\times 1}$) share \emph{all} the marginal and conditional distributions (\ie $p_{\thetav^{*}}(\xv_{\Sbb})=q(\xv_{\Sbb})$ and $p_{\thetav^{*}}(\xv_{\Tbb}|\xv_{\Sbb})=q(\xv_{\Tbb}|\xv_{\Sbb})$ hold for \emph{any non-overlapping} subsets $(\Sbb,\Tbb)$ of the index set $\Lbb=\{1,\cdots,L\}$); and
	
	\item both sides are identical in \emph{any} transformed domain, \ie with $\Xv=g(\xv)$ to transform $p_{\thetav^{*}}(\xv)/q(\xv)$ into $p_{\thetav^{*}}(\Xv)/q(\Xv)$ in the transformed $\Xv$ domain, we have $p_{\thetav^{*}}(\Xv_{\Tbb}|\Xv_{\Sbb})=q(\Xv_{\Tbb}|\Xv_{\Sbb}), \forall (\Sbb,\Tbb)$.
\end{itemize}

Below we first reveal that a single data sample demonstrates versatile data-sampling capabilities and that existing foundation models leverage a portion of these demonstrations to form their training/pretraining.
Next, based on the revelations, we propose our big cooperative learning (\textit{abbr.} big learning), which exhaustively exploits data information via versatile data-sampling demonstrations in a cooperative manner.
We then design tailored $2$-D simulations to explicitly demonstrate the principle of big cooperative learning, with additional interesting side-products.
Finally, we bring to light that big learning is a new dimension for upgrading conventional machine learning paradigms and, as an illustrative example, we upgrade the standard GAN into the big-learned BigLearn-GAN.

\subsection{Versatile but Underutilized Data-Sampling Demonstrations Within a Single Data Sample}
\label{sec:data_usage}

\begin{figure}%[H]
	%	\vspace{-5mm}
	\centering
	\subfloat[Example demonstrations of $\xv_{\Tbb} \sim q(\xv_{\Tbb}|\xv_{\Sbb}), \forall (\Sbb,\Tbb)$.]{
		\includegraphics[height=0.33\columnwidth]{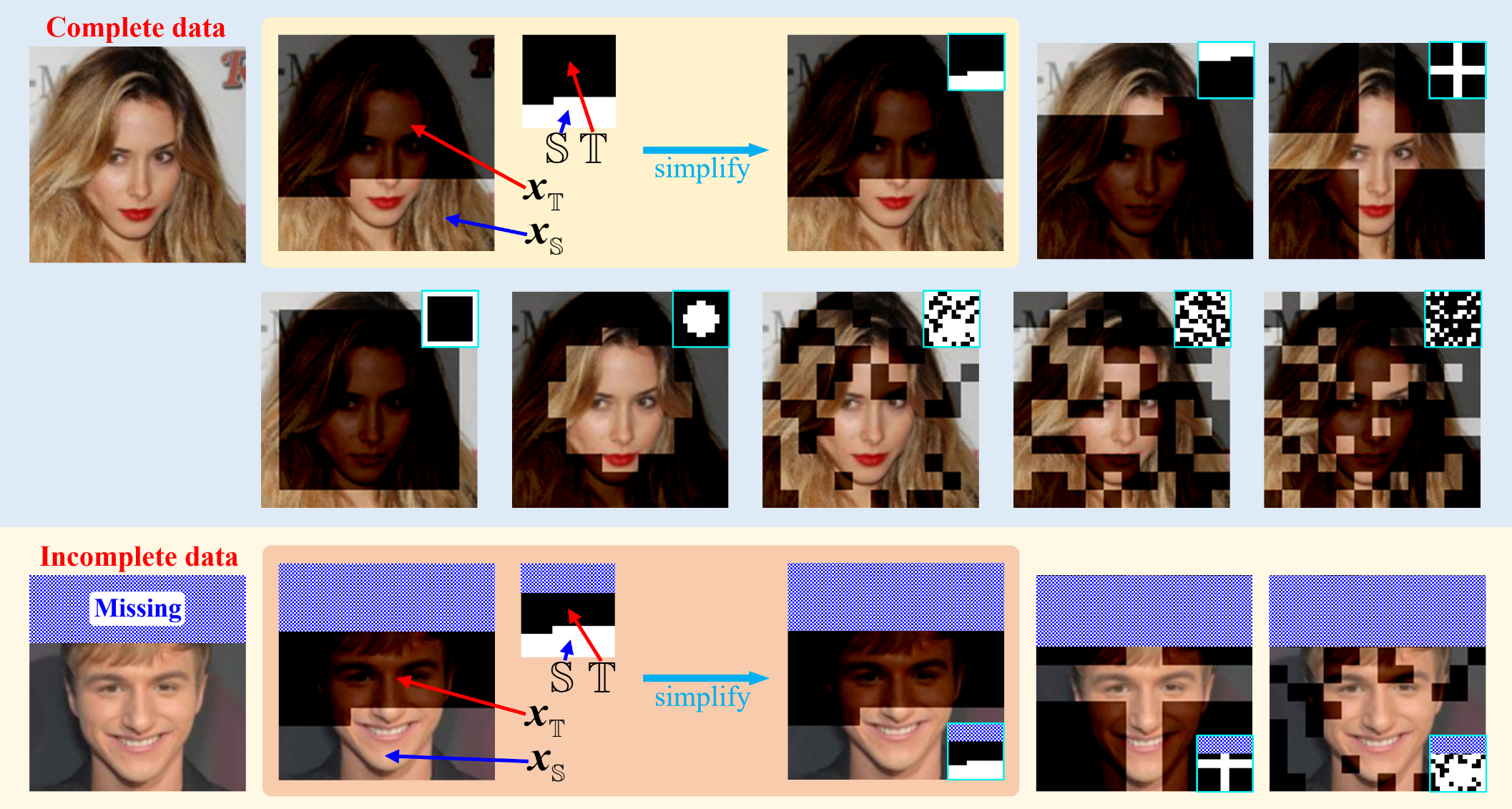}
		\label{fig:data_multi_perspective}}
	\quad
	\subfloat[$\Xv_{\Tbb} \sim q(\Xv_{\Tbb}|\Xv_{\Sbb}), \forall (\Sbb,\Tbb)$]{
		\includegraphics[height=0.33\columnwidth]{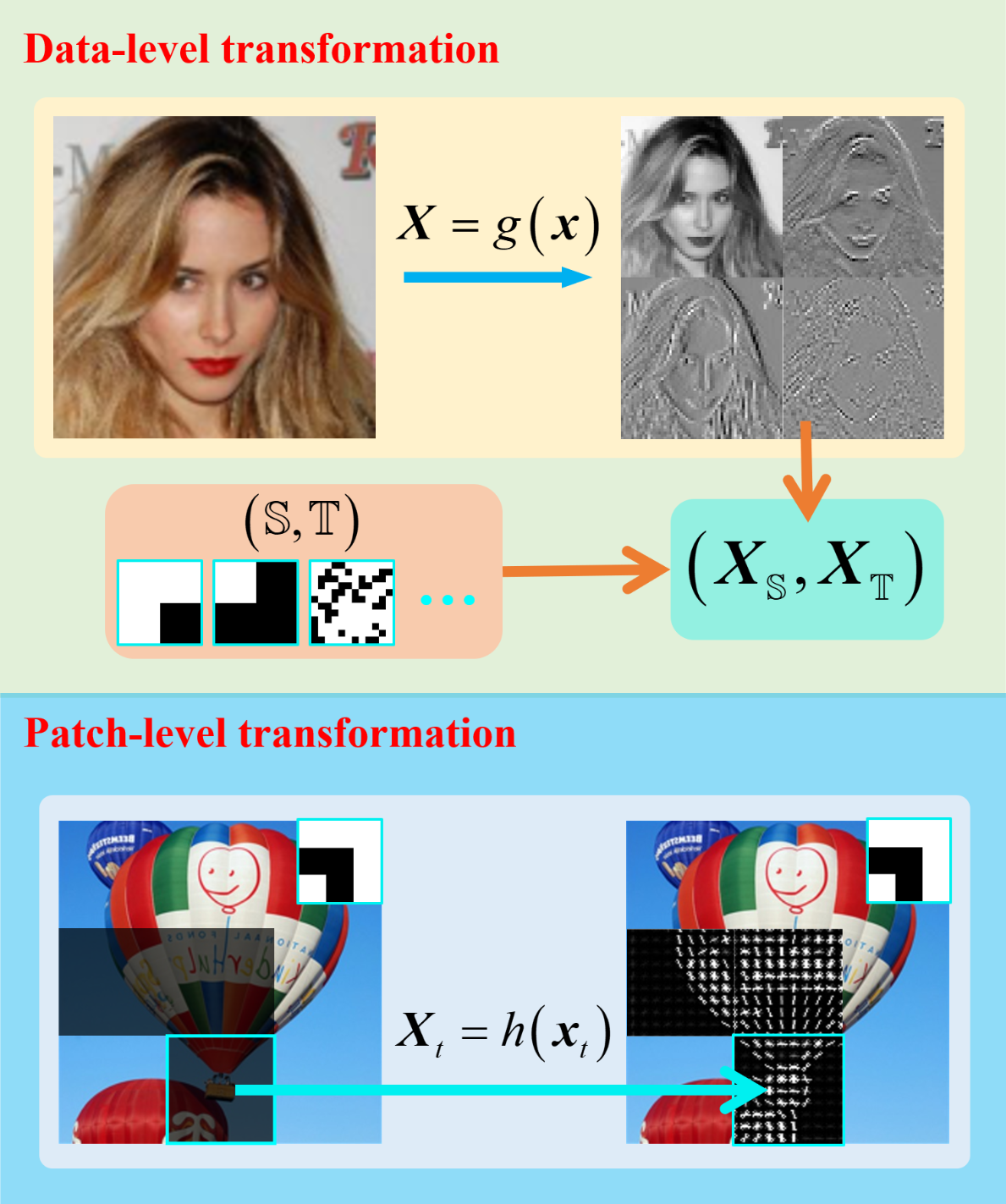}
		\label{fig:data_multi_transforms}}
	\caption{
		A single data sample demonstrates versatile data-sampling capabilities in the original domain (a) and diverse transformed domains (b).
		(a) Given a \emph{complete/incomplete} data sample, one simultaneously receives a demonstration for each $\xv_{\Tbb} \sim q(\xv_{\Tbb}|\xv_{\Sbb}), \forall (\Sbb,\Tbb)$.
		(b) Similarly, various data-sampling demonstrations across \emph{plentiful} transformed domains (\eg via a data-level transformation $g(\cdot)$ or a patch-level $h(\cdot)$) are also ready for exploitation.
	}
	\label{fig:data_multi}
	%	\vspace{-2mm}
\end{figure}

In conventional machine learning paradigms, a complete data sample $\xv \sim q(\xv)$ is often \emph{only} utilized in the \emph{joint} space (\ie all tokens $\{\xv_t\}_{t=1}^L$ of the joint $\xv=[\xv_1,\cdots,\xv_L]^T$ are always used simultaneously; see Eqs. \eqref{eq:mle} and \eqref{eq:standard_GAN_loss} for example).
However, we reveal in Fig. \ref{fig:data_multi} that, when given a single data sample (even it's incomplete), one simultaneously receives versatile data-sampling demonstrations (\ie samples from all the joint, marginal, and conditional data distributions; one sample per distribution) across potentially diverse domains.

These versatile data-sampling demonstrations, when accumulated across all data samples, actually constitute diverse datasets representing $q(\xv_{\Tbb}|\xv_{\Sbb})=p_{\thetav^{*}}(\xv_{\Tbb}|\xv_{\Sbb})$ or $q(\Xv_{\Tbb}|\Xv_{\Sbb})=p_{\thetav^{*}}(\Xv_{\Tbb}|\Xv_{\Sbb})$, $\forall (\Sbb,\Tbb)$, which are versatile outward manifestations of the data essence summarized in $\thetav^{*}$.
Before introducing how our big cooperative learning exploits such data-sampling demonstrations and leveraging tailored simulations to explicitly demonstrate their cooperativeness, we first reveal that existing foundation models use specific parts of these demonstrations to form their training.

Specifically, the mask-and-predict training uses a specific set of demonstrations from $q(\xv_{\Sbb^{\complement}}|\xv_{\Sbb})$, where, \eg $\Sbb$ is a random $85\%$ subset of $\Lbb$ in the BERT \cite{devlin2018bert}, to pretrain the conditionally-independent model $p_{\thetav}^{\text{MAE}}(\xv_{\Sbb^{\complement}}|\xv_{\Sbb})$ via minimizing the KL divergence $\Ebb_{q(\xv_{\Sbb})} \KL[q(\xv_{\Sbb^{\complement}}|\xv_{\Sbb}) || p^{\text{MAE}}_{\thetav}(\xv_{\Sbb^{\complement}}|\xv_{\Sbb})]$ (see Eqs. \eqref{eq:mask-and-predict} and \eqref{eq:mle}).
However, the trained model capabilities, \eg to independently predict the masked $15\%$ tokens conditioned on the unmasked $85\%$ tokens, are often discrepant from those required during the test phase, leading to the well-known problem of training-test (or pretraining-finetuning) discrepancy \cite{yang2019xlnet,yuan2022roadmap}.
On the other hand, the next-token-prediction training uses another specific set of data-sampling demonstrations from $q(x_{t}|\xv_{{<t}})$ to pretrain the auto-regressive model $p^{\text{AR}}_{\thetav}(x_{t}|\xv_{{<t}})$ by minimizing $\Ebb_{q(\xv_{{<t}})} \KL[q(x_{t}|\xv_{{<t}}) || p^{\text{AR}}_{\thetav}(x_{t}|\xv_{{<t}})]$ (see Eqs. \eqref{eq:next-token-prediction} and \eqref{eq:mle}).
Fortunately, the trained model capabilities of various next-token predictions often align closely with those required during testing (considering current applications), which results in a significantly reduced training-test discrepancy than that of mask-and-predict and is likely a key factor contributing to the success of large language models \cite{ChatGPT,GPT4,reid2024gemini,claude3}.
For better comparison, we summarize the utilized data-sampling demonstrations for both mask-and-predict and next-token-prediction, as well as other representative training methods for foundation models, in a consistent manner in Table \ref{tab:BigLearn_special_cases}.

\begin{table}%[H]
	%	\vspace{-3mm}
	\centering
	\caption{Big cooperative learning and representative training objectives of foundation models, where $\xv\in \Rbb^{L\times D}\sim q(\xv)$, $\Lbb=\{1,\cdots,L\}$, $\Sbb \subset \Lbb$, $\Tbb \subseteq \Lbb, \Tbb \neq \emptyset$, and $\Sbb\cap\Tbb = \emptyset$.
	}
	\resizebox{0.95\columnwidth}{!}{
		\begin{tabular}{l l l}
			\hline\hline 
			$\quad$ Training Method & $\qquad\qquad$ Training Objective & $\qquad\qquad\quad$Settings \\
			\hline\hline
			\multirow{3}{*}{Big Cooperative Learning} & \multirow{3}{*}{
				\makecell[l]{$p_{\thetav}(\Xv_{\Tbb}|\Xv_{\Sbb}) \longrightarrow q(\Xv_{\Tbb}|\Xv_{\Sbb}), \forall ({\Sbb},{\Tbb})\in \Omegamat$}
			} & 
			\multirow{3}{*}{
				\makecell[l]{$\longrightarrow$: a divergence/distance of PDFs\\$\Omegamat$: a user-defined set of $(\Sbb,\Tbb)$s\\$\Xv$: transformed $\xv$ (see Fig. \ref{fig:data_multi_transforms})}
			} \\
			& & \\
			& & \\
			\hdashline
			\multirow{3}{*}{BigLearn-GAN} & \multirow{3}{*}{
				\makecell[l]{$\Ebb_{q(\Sbb,\Tbb)q(\Xv_{\Sbb})} \JS[q(\Xv_{\Tbb}|\Xv_{\Sbb})||p_{\thetav}(\Xv_{\Tbb} | \Xv_{\Sbb})]$
					%$\Ebb_{q(\Sbb, \Tbb)} \JS[q(\xv_{\Sbb\cup\Tbb})||p_{\thetav}(\xv_{\Tbb} | \xv_{\Sbb}) q(\xv_{\Sbb})]$ %+ \\$\Ebb_{q(\Sbb^1, \Tbb^1)q(\Sbb^2, \Tbb^2)} \JS[p_{\thetav}(\xv_{\Tbb^1} | \xv_{\Sbb^1})q(\xv_{\Sbb^1})||p_{\thetav}(\xv_{\Tbb^2} | \xv_{\Sbb^2}) q(\xv_{\Sbb^2})]$
			}} & 
			\multirow{3}{*}{\makecell[l]{
					$q(\Sbb, \Tbb)$: PDF of $(\Sbb, \Tbb)$\\
					$p_{\thetav}(\Xv_{\Tbb} | \Xv_{\Sbb})$: generative processes
					% implicitly modeled as
					% $(\Sbb, \Tbb)/(\Sbb^1, \Tbb^1)/(\Sbb^2, \Tbb^2)$: \iid  \\\qquad\quad samples from $q(\Sbb, \Tbb)$
			}} \\
			& & \\
			& & \\
			\hline\hline
			\multirow{3}{*}{Mask-and-predict} & \multirow{3}{*}{
				$\Ebb_{q(\Sbb)q(\xv_{\Sbb})} \KL[q(\Xv_{\Sbb^{\complement}}|\xv_{\Sbb})||p_{\thetav}^{\text{MAE}}(\Xv_{\Sbb^{\complement}}|\xv_{\Sbb})]$
			} & 
			\multirow{3}{*}{\makecell[l]{
					$\Tbb=\Sbb^{\complement}$, $\Xv_{\Tbb}=\xv_{\Tbb}$ in BERT \cite{stickland2019bert}\\
					$\Xv_{\Tbb}$ is a normalized $\xv_{\Tbb}$ in MAE \cite{he2021masked}\\
					$\Xv_{\Tbb}$ is a HOG-transformed $\xv_{\Tbb}$ in \cite{wei2021masked}
			}} \\
			& & \\
			& & \\
			\hline\hline
			\multirow{3}{*}{Next-token-prediction} & \multirow{3}{*}{
				$\Ebb_{q(t)q(\xv_{{<t}})} \KL[q(x_{t}|\xv_{{<t}}) || p^{\text{AR}}_{\thetav}(x_{t}|\xv_{{<t}})]$
			} & 
			\multirow{3}{*}{\makecell[l]{
					$q(t) = U[1,L]$, $\Sbb=\{<t\}$, $\Tbb=\{t\}$\\Used in large language models \cite{ChatGPT,GPT4}
			}} \\
			& & \\
			& & \\			
			\hline\hline
			\multirow{3}{*}{\makecell[l]{Permutation language\\modeling \cite{yang2019xlnet}}} & 
			\multirow{3}{*}{
				$\Ebb_{q(\zv)q(t)q(\xv_{\zv_{<t}})} \KL[q(x_{z_{t}}|\xv_{\zv_{<t}}) || p^{\text{AR}}_{\thetav}(x_{z_t}|\xv_{\zv_{<t}})]$
			} & 
			\multirow{3}{*}{\makecell[l]{
					$q(\zv)$: PDF of a random permutation $\zv$\\
					$\Sbb=\{\zv_{<t}\}$, $\Tbb=\{z_t\}$
			}} \\
			& & \\
			& & \\
			\hline\hline
		\end{tabular}
	}
	\label{tab:BigLearn_special_cases}
	%		\vspace{-2mm}
\end{table}

Considering various testing scenarios for foundation models, especially those associated with artificial general intelligence, severe training-test discrepancy may arise from training with a limited portion of the \emph{available but underutilized} data-sampling demonstrations, pinpointing the insufficiency of the existing training theory for foundation models \cite{bommasani2021opportunities}.
We propose to exhaustively exploit these demonstrations during training to significantly expand the training scope to reduce the training-test discrepancy.

\subsection{Big Cooperative Learning for Exhaustive Data Exploitation}
\label{sec:big_learning}

Our main motivations include 
$(i)$ versatile data-sampling demonstrations are readily given in data samples (see Fig. \ref{fig:data_multi}),
$(ii)$ these demonstrations can be used to form massive learning tasks (\ie \textbf{learning individuals} in human cooperative learning),
$(iii)$ they are different manifestations of the \emph{same and unique} data essence that is summarized in $\thetav^{*}$ (\ie \textbf{the common goal}; refer to the above \textbf{Assumptions}),
and $(iv)$ more importantly, their corresponding data-sampling capabilities are likely to be those required when considering general test scenarios.
Accordingly, we propose to maximally reduce the training-test discrepancy with the presented big cooperative learning (\textit{abbr.} big learning), which exhaustively exploits the available data-sampling demonstrations to construct massive diverse learning tasks/individuals that \emph{cooperatively} approach the common goal, as detailed below.

\begin{definition}[Big cooperative learning]
	\label{def:biglearn}
	Based on the \textbf{Assumptions} in Section \ref{sec:our_method} and given data samples $\xv\in \Rbb^{L\times D}$ (with length $L$ and dimension $D$) from the underlying data distribution $q(\xv)$ and a universal model $p_{\thetav}(\cdot)$ with sufficient modeling capacity,
	\textbf{big cooperative learning} ({abbr.} big learning) trains the universal parameter set $\thetav$ in a massively multitasking cooperative manner as 
	\beq\bali\label{eq:big_learning}
		p_{\thetav}(\Xv_{\Tbb}|\Xv_{\Sbb}) \longrightarrow q(\Xv_{\Tbb}|\Xv_{\Sbb}), \forall ({\Sbb},{\Tbb})\in \Omegamat,
	\eali\eeq
	where the arrow ``$\rightarrow$'' means using its LHS distribution to match its RHS one via some divergence/distance metric, $\Xv\in \Rbb^{L\times D}$\footnote{
		In general, the shapes of $\xv$ and $\Xv$ can be different.
		We assume they are the same for simplified notations.
	} stands for transformed $\xv$ in potentially many transformed domains, $\Sbb$ and $\Tbb$ (randomly sampled from $\Omegamat$) are {non-overlapping} subsets of the length index set $\Lbb=\{1,\cdots,L\}$, and $q(\Xv_{\Tbb}|\Xv_{\Sbb})$ with different $(\Sbb,\Tbb)$s can represent a joint/marginal/conditional distribution in a transformed domain, whose samples are readily available/transformed from $\{\xv\}$.
\end{definition}

\begin{remark}
	The divergence/distance metric for the arrow ``$\rightarrow$'' is application dependent; for example, in situations where maximum-likelihood learning is popular, ``$\rightarrow$'' often represents the forward KL divergence.
	Often metrics of the same category are employed for various $(\Sbb,\Tbb)$ pairs.
%	; whereas for GAN-related situations, ``$\rightarrow$'' may represent the JS divergence.
\end{remark}

\begin{remark}\label{remark:diverse_transform}
	Data-level transformations $\Xv = g(\xv)$ (like noising) and/or patch-level ones $\Xv_{t} = h(\xv_{t})$ (see Fig. \ref{fig:data_multi_transforms}) can be employed simultaneously to construct matchings across diverse transformed domains.
	Such transformations can be embedded in the modeling of $p_{\thetav}(\cdot)$.
	Often $\Xv = \xv$.
\end{remark}

\begin{remark}
	$\Omegamat$ consists of $(\Sbb,\Tbb)$ pairs of interest, where $\Sbb \subset \Lbb$ and $\Tbb \subseteq \Lbb, \Tbb \neq \emptyset$.
	Often a distribution $q(\Sbb,\Tbb)$ is defined for sampling $(\Sbb,\Tbb)$ from $\Omegamat$.
	Note that $\Sbb \cup \Tbb$ need not be $\Lbb$, meaning that incomplete data are naturally utilized in the corresponding marginal/conditional matching tasks.
\end{remark}

\begin{remark}
	To address the modeling challenge arising from versatile settings of $(\Xv_{\Sbb},\Xv_{\Tbb})$, the flexible transformer architecture is often employed to construct the universal model $p_{\thetav}(\Xv_{\Tbb}|\Xv_{\Sbb})$.
\end{remark}

\begin{remark}\label{remark:multimodal_biglearn}
	Big cooperative learning can be generalized to handle multi-modal applications with paired data $\xv=[\xv',\yv',\zv',\cdots]$, where data incompleteness may have an additional modality-level hierarchy. Similarly, the employed transformations for $\Xv$, the metric for the arrow ``$\rightarrow$'', and the modeling of $p_{\thetav}(\cdot)$ should be further developed to address the challenges posed by multiple data modalities.
	Appendix \ref{appsec:big_learning_multimodal} provides an initial discussion.
\end{remark}

Big cooperative learning in Definition \ref{def:biglearn} exhaustively exploits the available data-sampling demonstrations inherent in the data to form massive learning tasks that \emph{cooperatively} train the universal $p_{\thetav}(\cdot)$ towards the common goal (\ie the unique data essence summarized in $\thetav^{*}$).
Below we design tailored $2$-D simulations to explicitly demonstrate the principle of big cooperative learning.

\subsection{Tailored $2$-D Simulations to Demonstrate the Big-Learning Principle}
\label{sec:2D_gmm_simulation}

It's not easy to comprehend the principle of a general concept like big cooperative learning. To explicitly demonstrate the cooperation among the diverse matching tasks in Eq. \ref{eq:big_learning}, we leverage Gaussian Mixture Models (GMMs) to design simple but controllable $2$-D simulations, where 
$(i)$ the model capacity is guaranteed to be sufficient and
$(ii)$ the loss of each matching task can be computed traversely and shown as an image (thus no need to consider the influence of optimization).

Specifically, both $q(\xv)$ and $p_{\thetav}(\xv)$ are set as a GMM with $K=2$ components (\emph{abbr.} $2$-GMM), \ie
\beq
	q(\xv) = p_{\thetav^{*}}(\xv) = \sum\nolimits_{i=1}^2 \frac{1}{2} \Nc(\xv|\muv^{*}_i, \Sigmamat^{*}_i)
	\quad \text{and} \quad
	p_{\thetav}(\xv) = \sum\nolimits_{i=1}^2 \frac{1}{2} \Nc(\xv|\muv_i, \Sigmamat_i),
\eeq
where $\muv^{*}_1 = [-1, 0]^T$, $\muv^{*}_2 = [1, 0]^T$, $\muv_1 = [\mu_1, 0]^T$, $\muv_2 = [\mu_2, 0]^T$, $\thetav=[\mu_1,\mu_2]^T$ contains $2$-D trainable parameters, and $\Sigmamat^{*}_1 = \Sigmamat^{*}_2 = \Sigmamat_1 = \Sigmamat_2 = \sigma^2 \Imat$ with hyperparameter $\sigma^2$.
Accordingly, Eq. \ref{eq:big_learning} with $\Xv=\xv$ consists of $1$ joint matching, $2$ marginal matchings, and $2$ types of conditional matchings.
We set the arrow ``$\rightarrow$'' as the \emph{reverse} KL divergence, which is closely related to adversarial learning. For example, the loss for joint matching is $\KL[p_{\thetav}(\xv)||q(\xv)]$, which is a function \wrt the $2$-D $\thetav$.

\begin{figure*}%[H]
	%	\vspace{-5mm}
	\centering
	\includegraphics[width=\columnwidth]{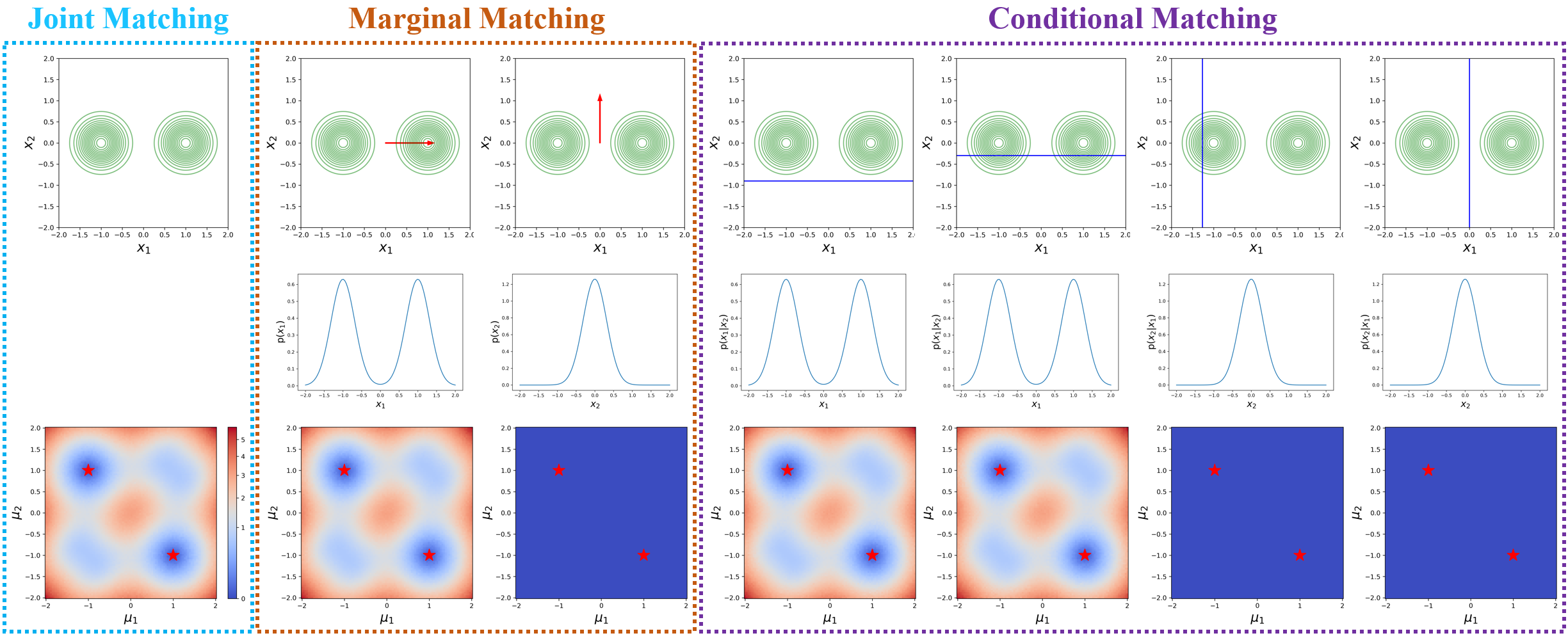}
	\vspace{-3mm}
	\caption{Demonstrations of the data distribution and the reverse-KL loss surfaces for for joint, marginal, and conditional matchings in the tailored $2$-D simulations.
		The first row illustrates the joint distribution $q(\xv)$ and the marginal/conditional space of interest.
		The second row shows the corresponding marginal/conditional data distribution.
		The last row exhibits the surface of $\KL[p_{\thetav}(\xv)||q(\xv)]$, $\KL[p_{\thetav}(x_i)||q(x_i)], i\in\{1,2\}$, and $\KL[p_{\thetav}(x_i|x_j)||q(x_i|x_j)], j\neq i$, respectively.
		The two global optima are marked with red stars. $\sigma^2=0.1$. 
	}
	\label{fig:naive_JMC_not_work}
	\vspace{-2mm}
\end{figure*}

Fig. \ref{fig:naive_JMC_not_work} demonstrates the joint, marginal, and conditional matchings in the original $\xv$-domain, where 
$(i)$ naive joint matching has a loss with significant local minima and 
$(ii)$ in situations with independent features/tokens (rare in practice), there is no space for joint, marginal, and conditional matchings to help (or cooperate with) each other in the original space.

\begin{figure*}%[H]
	%	\vspace{-5mm}
	\centering
	\includegraphics[width=\columnwidth]{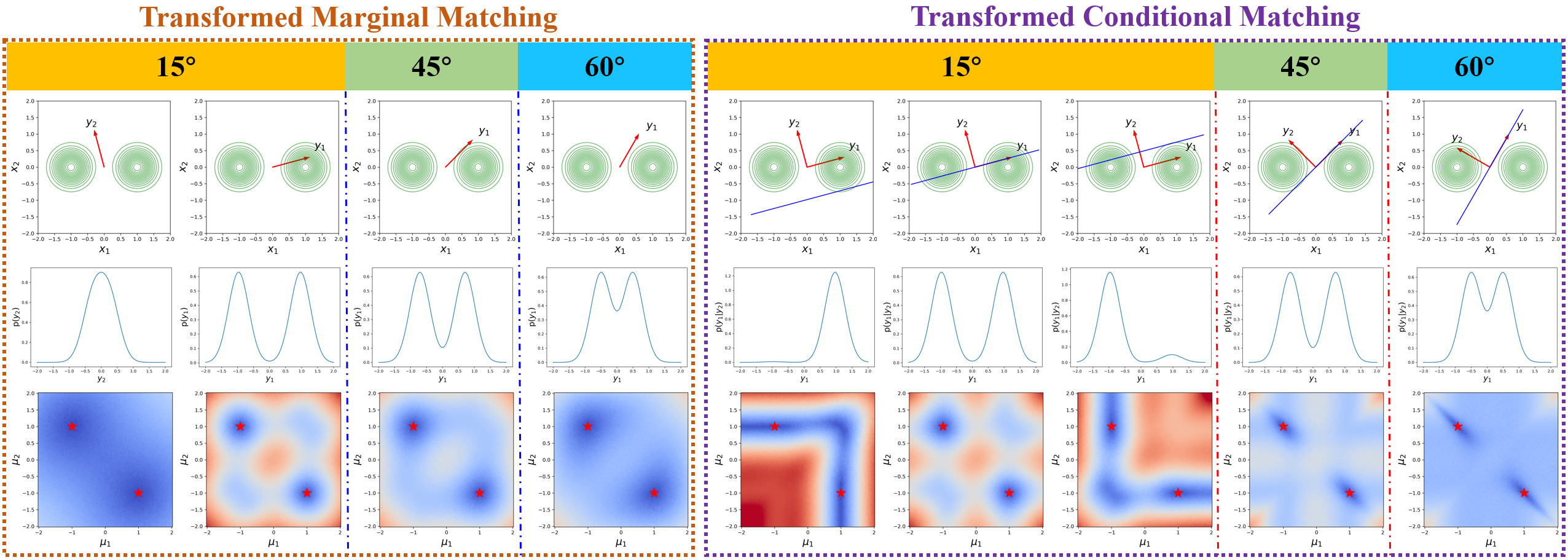}
	\vspace{-3mm}
	\caption{Demonstrations of marginal and conditional matchings in diverse rotationally transformed domains. 
		The joint matching remains the same as in Fig. \ref{fig:naive_JMC_not_work} after a rotation transformation. 
		$\sigma^2=0.1$. 
		The local optima unstably vary with different matchings but the global optima are stably the same, laying the foundation for the cooperation among diverse matchings as in Eq. \eqref{eq:big_learning}.
	}
	\label{fig:trans_MC_matching}
	%	\vspace{-2mm}
\end{figure*}

Considering that practical applications often have dependent tokens (\eg correlated image patches), we reveal that \textbf{the dependence among tokens enables cooperation} among joint, marginal, and conditional matchings.
Specifically, we modify the above $2$-D simulation by using a rotation transformation (\ie $\yv=\Amat\xv$) to introduce dependence among tokens of the transformed $\yv$; we then perform joint, marginal, and conditional matchings in the transformed $\yv$-domain. 
Fig. \ref{fig:trans_MC_matching} demonstrates the loss surfaces when $\Amat$ represents $15^{\circ}$, $45^{\circ}$, and $60^{\circ}$ rotations, respectively.
It's evident from the losses associated with the $15^{\circ}$ rotation that, in situations with dependent tokens, \textbf{cooperation arise among different matchings because local optima unstably vary with the matching task but the global optima are stably the same};
similar phenomena also hold across diverse transformed domains (\eg see the loss surfaces of the $y_1$-marginal matchings under different rotations).
Therefore, multitask learning with diverse matchings is expected to overlook the unstable local optima but focus on the stable global ones, manifested as a power of exploration.

The explicit demonstrations in Fig. \ref{fig:trans_MC_matching} justify the principle of our big cooperative learning (with Eq. \eqref{eq:big_learning}) in a lightweight way, where different learning individuals (\ie matching tasks with different local optima) \emph{cooperatively} approach the common goal (\ie the global optima representing the data essence).
Furthermore, they also serve as a lightweight learning-perspective explanation for the great successes of foundation models.

During our investigations in the tailored $2$-D simulations, we discover several interesting side-products that potentially benefit implementations of improved foundation models.
Due to space limitations, we summarize them below and give detailed discussions in Appendix \ref{appsec:tailored_2D_simulation}.
\begin{itemize}[leftmargin=5mm]	
	\item Big learning constructed with diverse transformed joint and marginal matchings may favor bi-level optimization (\ie training multiple steps in one matching before moving on to the next).
	
	\item Big learning constructed with diverse transformed conditional matchings may straightforwardly overlook local optima and need no bi-level optimization.
	
	\item Multi-scale noising (when applicable) serves as powerful data-level transformations for big cooperative learning; this is akin to diffusion models \cite{ho2020denoising,song2020score}. 
	
\end{itemize}

\subsection{Big Learning as a New Dimension for Upgrading Machine Learning Paradigms}
\label{sec:BL_upgrade_dimension}

After noticing that 
$(i)$ groundbreaking foundation models are pretrained by exploiting a potion of data-sampling demonstrations inherent in their training data,
$(ii)$ these demonstrations can be exhaustively exploited to form versatile matchings (see Eq. \eqref{eq:big_learning}), among which cooperation could arise to help overlook local optima and focus on global ones,
$(iii)$ the data-sampling demonstrations are also available in conventional machine learning paradiagms, where often only joint matching is employed (see Eqs. \eqref{eq:mle} and \eqref{eq:standard_GAN_loss}),
we bring to light that big cooperative learning is a new dimension for upgrading conventional machine learning paradigms.

Before presenting an illustrative example, we first note that a similar idea already underlies several recent/concurrent papers.
For example, \cite{cong2024big} simultaneously employs joint, marginal, and transformed marginal matchings to effectively address the local-optimum challenge of the conventional expectation-maximization algorithm.
\cite{bao2023one} trains a diffusion model via simultaneous joint, marginal, and conditional matchings and delivers diverse SOTA cross-modality data-sampling capabilities.

Below we leverage our big cooperative learning to upgrade the standard GAN \cite{goodfellow2014generative} into its big-learning variant, termed BigLearn-GAN, which can alternatively be interpreted as a novel adversarial pretraining method for foundation models. 
Specifically, given data $\xv$ and the transformed \emph{continuous} $\Xmat\in \Rbb^{L\times D}$, the BigLearn-GAN performs big cooperative learning in Definition \ref{def:biglearn} (see Eq. \eqref{eq:big_learning}) in an adversarial-learning fashion with the objective of 
\beq\label{eq:model_to_data_all}
	\min_{\thetav} \max_{\phiv} 
	\Ebb_{q(\Sbb,\Tbb)q(\Xv_{\Sbb})}\big[
	\Ebb_{q(\Xv_{\Tbb}|\Xv_{\Sbb})} {\log} \sigma[f_{\phiv}(\Xv;\Sbb,\Tbb)] + 
	\Ebb_{p_{\thetav}(\Xv_{\Tbb} | \Xv_{\Sbb})} {\log} \sigma[-f_{\phiv}(\Xv;\Sbb,\Tbb)]
	\big],
\eeq
where the universal $p_{\thetav}(\Xv_{\Tbb} | \Xv_{\Sbb})$ models the {generative processes} of outputs $\Xv_{\Tbb}$ conditioned on inputs $\Xv_{\Sbb}$ for \emph{all} $(\Sbb,\Tbb)$ pairs (an example implementation based on the transformer architecture is given in Appendix \ref{appsec:BigLearn_GAN_archtecture}).
Similar to the standard GAN \cite{goodfellow2014generative}, with optimal discriminators where $f_{\phi^{*}}(\Xv;\Sbb,\Tbb) = {\log} \frac{q(\xv_{\Tbb} | \xv_{\Sbb})}{p_{\thetav}(\xv_{\Tbb} | \xv_{\Sbb})}$, Eq. \eqref{eq:model_to_data_all} is equivalent to $\min_{\thetav} \Ebb_{q(\Sbb,\Tbb)q(\Xv_{\Sbb})} \JS[q(\Xv_{\Tbb}|\Xv_{\Sbb})||p_{\thetav}(\Xv_{\Tbb} | \Xv_{\Sbb})]$ \cite{goodfellow2014generative}.
Note that with different settings of $(\Sbb,\Tbb)$, $q(\Xv_{\Tbb}|\Xv_{\Sbb})$ can recover a joint/marginal/conditional distribution; accordingly, the user-defined $q(\Sbb,\Tbb)$ determines the weighting among different matchings.

Details are given in Appendix \ref{appsec:BigLearn_GAN}. After big cooperative learning, the BigLearn-GAN is expected to deliver a universal $p_{\thetav}(\Xv_{\Tbb} | \Xv_{\Sbb}), \forall (\Sbb,\Tbb)$ that simultaneously possesses versatile synthesis capabilities across potentially diverse domains, therefore demonstrating generalizability to a wide range of test situations requiring these capabilities.

\section{Experiments}
\label{sec:Exp}
%\vspace{-2mm}

%Exp
%RKL GMM exp to demonstrate the strong exploration power from BL
%
%BLGAN versatile completion capabilities
%
%multimodal application on finetuning and class-gen

We design challenging simulations based on the tailored ones in Section \ref{sec:2D_gmm_simulation} and conduct several high-dimensional experiments to further demonstrate the principle of big cooperative learning, which, in turn, also justifies the successes of foundation models from the perspective of learning.

Specifically, we first design a $25$-GMM reverse-KL-minimization simulation, where severe mode collapse emerges from numerous local optima for conventional joint reverse-KL matching; however, our empirical findings demonstrate that big learning is capable of delivering remarkable exploration power.
Next, we implement our BigLearn-GAN on the uni-modal MNIST and CelebA datasets and demonstrate the versatile realistic data-sampling capabilities of the big-learned universal model.
Finally, we expose the potential of big learning in multi-modal applications, \eg to unify classification and generation or to serve as a superior fine-tuning strategy.

\subsection{Remarkable Exploration Power of Big Cooperative Learning}
\label{sec:25GMM_exploration}

\begin{figure}[tb]
	%	\vspace{-3mm}
	\centering
	\includegraphics[width=\columnwidth]{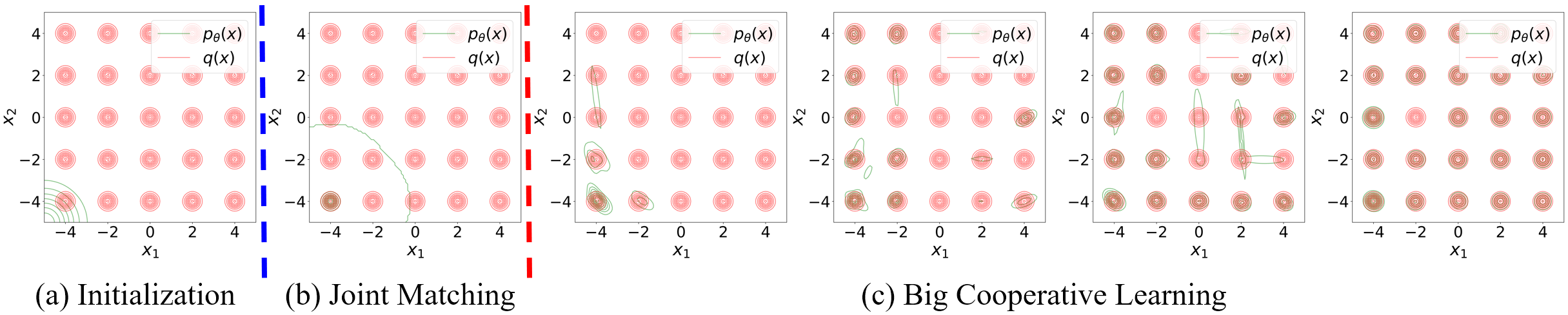}
	\vspace{-3mm}
	\caption{Illustrating the exploration power of big learning on the $25$-GMM reverse-KL-minimization simulation.
		(a) The challenging initialization.
		(b) Joint matching gets stuck in local optima.
		(c) Big learning gradually seeks out most components; results of $200$, $800$, $1400$, $6000$ iterations are shown.
	}
	\label{fig:25GMM_Exploration}
\end{figure}

Based on Section \ref{sec:2D_gmm_simulation}, we conduct a more challenging reverse-KL-minimization simulation, where both $q(\xv)$ and $p_{\thetav}(\xv)$ are set as a $25$-GMM and $\xv\in\Rbb^2$ (see Fig. \ref{fig:25GMM_Exploration} and Appendix \ref{appsec:25GMM_exploration} for details). 
The loss for conventional joint matching, \ie $\KL[p_{\thetav}(\xv)||q(\xv)]$, has numerous local optima (refer to Fig. \ref{fig:naive_JMC_not_work}), which are closely related to the notorious mode collapse problem of adversarial learning.

To demonstrate the remarkable exploration power of big learning, we select a challenging initialization, where all the $25$ components of $p_{\thetav}(\xv)$ are initialized in the lower-left corner of Fig. \ref{fig:25GMM_Exploration}a.
As expected, conventional joint matching quickly gets stuck in a strong local optimum, as shown in Fig. \ref{fig:25GMM_Exploration}b.
In comparison, our big learning manages to deliver a remarkable exploration power (see Fig. \ref{fig:25GMM_Exploration}c), which overlooks local optima and gradually seeks out most components in the mode-seeking reverse-KL-minimization territory.
It's worth noting that more powerful exploration may be harvest from versatile matchings across diverse transformed domains, as discussed in Appendix \ref{appsec:25GMM_exploration}.

\subsection{Versatile Realistic Data-Sampling Capabilities of the Universal BigLearn-GAN Generator}
\label{sec:VersatileCapabilities_biglearning}

Next, we implement the BigLearn-GAN in Section \ref{sec:BL_upgrade_dimension} on the MNIST and CelebA datasets and demonstrate the big-learned versatile realistic data-sampling capabilities of the universal generator, providing empirical evidence that big cooperative learning is a new dimension for upgrading conventional machine learning paradigms. 
We employ $\Xv=\xv$ and the transformer architecture to design both the generator and discriminator in Eq. \eqref{eq:model_to_data_all}; further details are provided in Appendix \ref{appsec:BigLearn_GAN}.

\begin{figure}[tb]
	%	\vspace{-3mm}
	\centering
	\includegraphics[width=\columnwidth]{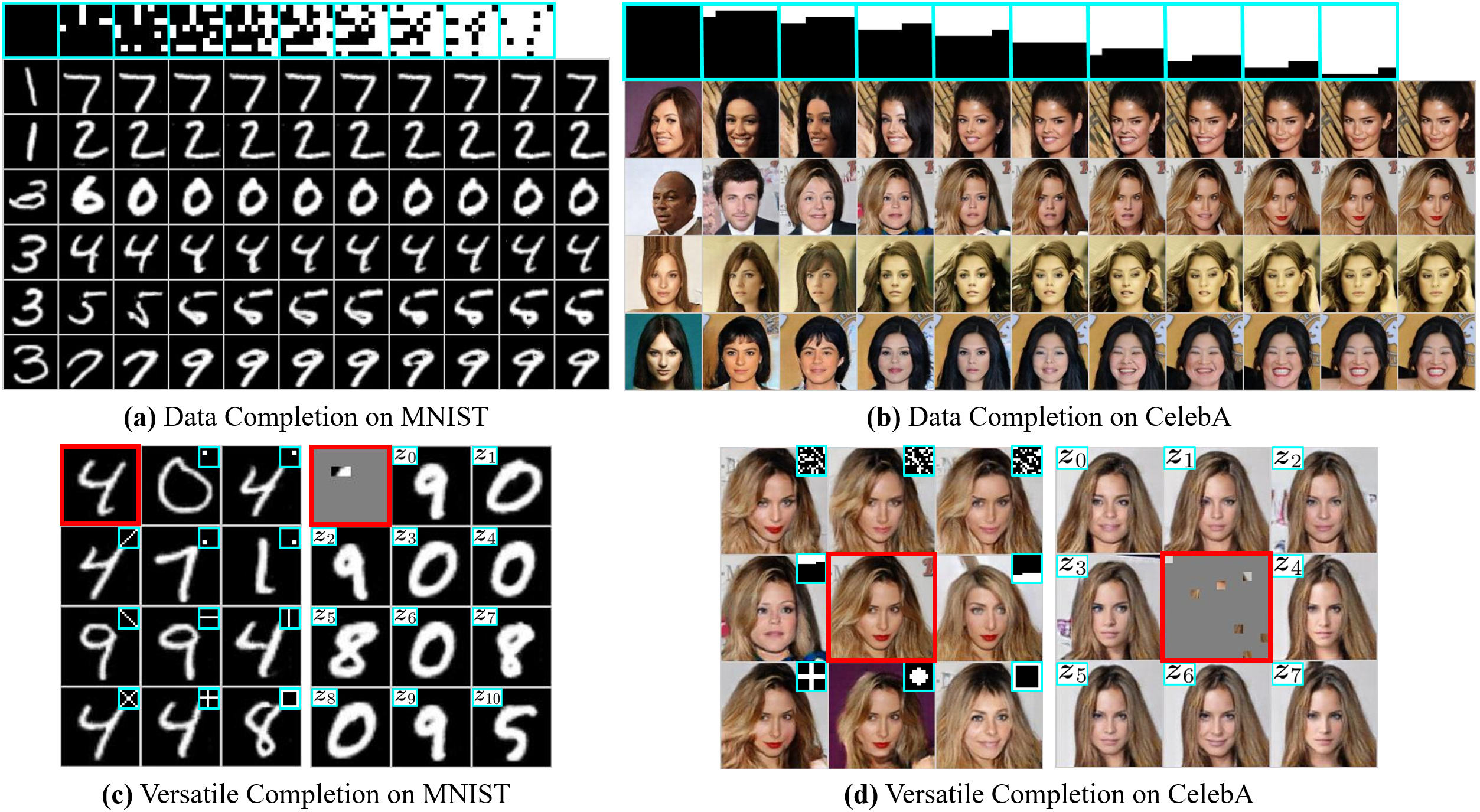}
	\vspace{-3mm}
	\caption{Versatile realistic data-sampling capabilities of the big-learned BigLearn-GAN. 
		(a-b) Data completion with random/initial-portion $\Sbb$, where $\Sbb$s are shown in the first row (light-blue boxed), real images $\xv$s are given in the rightmost column, generated images with $p_{\thetav}(\xv_{\Tbb}|\xv_{\Sbb})$ are shown in the rest rows and columns.
		(c-d) Versatile completion \wrt various $\Sbb$s (left) and \wrt various noise $\zv$s but the same $\xv_{\Sbb}$ (right). 
		The light-blue boxes show $\Sbb/\zv$s, while the red ones show $\xv$ (left) or $\xv_{\Sbb}$ (right). 
	}
	\label{fig:versetile_completion_BigLearnGAN}
	\vspace{-2mm}
\end{figure}

Demonstrations of versatile data-sampling capabilities are summarized in Fig. \ref{fig:versetile_completion_BigLearnGAN}, where the big-learned universal generator $p_{\thetav}(\xv_{\Tbb}|\xv_{\Sbb})$ simultaneously possesses realistic generation/completion power \wrt diverse settings of $\xv$, $\Sbb$, and noise $\zv$ (note the adaptive generation diversity \wrt different amount of information in $\xv_{\Sbb}$; see Appendix \ref{appsec:add_results_BigLearn-GAN} for more results).
This is expected because these data-sampling capabilities, \emph{being cooperative}, are explicitly trained.
On the other hand, such versatile data-sampling capabilities would also be valuable and generalize to a wide range of downstream testing applications, highlighting the effectiveness of big cooperative learning.

\subsection{Revealing the Potential of Big Cooperative Learning in Multi-Modal Applications}
\label{sec:MultiModal_BigLearn_exp}

Following Remark \ref{remark:multimodal_biglearn} of Definition \ref{def:biglearn}, we additionally test the presented big learning in general multi-modal settings.
Specifically, we consider $2$ simplified situations with two modalities (\ie $\xv = (\xv', y')$ contains paired tokens $\xv'$ and a label $y'$), the first of which unifies classification and generation on MNIST, while the other one concerns using big learning as a fine-tuning strategy on the GLUE benchmark \cite{wang2018glue}.
We use $\Xv=\xv$; more details are given in Appendix \ref{appsec:biglearn_MultiModal}.
After big cooperative learning, the universal $p_{\thetav}(\xv_{\Tbb}|\xv_{\Sbb})$ is expected to simultaneously deliver versatile capabilities by specifying different $(\Sbb, \Tbb)$s, such as generation $p_{\thetav}(\xv')$, conditioned generation $p_{\thetav}(\xv'|y')$, classification $p_{\thetav}(y'|\xv')$, various completion $p_{\thetav}(\xv'_{\Tbb}|\xv'_{\Sbb})$, conditioned completion $p_{\thetav}(\xv'_{\Tbb}|\xv'_{\Sbb}, y')$, \etc

Due to space limitations, we move the results to Appendix \ref{appsec:biglearn_MultiModal} and present a brief summary here. 
Similar to the versatile data-sampling capabilities demonstrated in Section \ref{sec:VersatileCapabilities_biglearning}, the former MNIST experiment proves that big cooperative learning is also capable of providing versatile \emph{cross-modal} capabilities with a universal model.
The latter GLUE experiments reveal that big learning has the potential to serve as a superior fine-tuning strategy than the naive one.

%\begin{figure}[tb]
%		\vspace{-2mm}
%	\centering
%	\subfloat[Joint Generation]{
%		\includegraphics[height=0.2\columnwidth]{Figures/MNISTgen.png}
%		\label{fig:}}
%	\qquad
%	\subfloat[Label-Conditioned Generation ]{
%		\includegraphics[height=0.2\columnwidth]{Figures/MNISTcondgen.png}
%		\label{fig:}}
%	\vspace{-2mm}
%	\caption{
%		Demonstration of versatile data capabilities of big learning, retrieved from $p_{\thetav}(\Xv_{\Tbb'}|\Xv_{\Sbb'})$ with specified $(\Sbb', \Tbb')$.
%	}
%	\label{fig:biglearn_genclass}
%		\vspace{-2mm}
%\end{figure}

%\begin{table}[tb]
%	\centering
%	\caption{Big learning serves as a superior fine-tuning strategy.
%		The best/median metrics are calculated among the combinations of the tested hyperparameters of Table \ref{apptab:glue_hyperpara}.}
%%	\vspace{2mm}
%		\resizebox{0.6\columnwidth}{!}{
%		\begin{tabular}{l | c c | c c}
%			\hline \hline
%			\multirow{2}{*}{Task} & \multicolumn{2}{c|}{Best Accuracy / F1} & \multicolumn{2}{c}{Median Accuracy / IQR}
%			\\ 
%			& FT & big-learn & FT & big-learn
%			\\ \hline 
%			RTE & 71.84  & $\textbf{75.09}$ & 66.06/2.34  & $\textbf{70.75/1.44}$ 
%			\\
%			MRPC & 88.97/92.09 & $\textbf{90.20/93.03}$ & 87.00/2.45 & $\textbf{87.74/1.10}$  
%			\\
%			SST-2 & 94.15  & $\textbf{95.18}$ & 93.75/0.45 & $\textbf{94.66/0.28}$  
%			\\ \hline \hline
%		\end{tabular}
%		}
%	\label{apptab:glue_ACC}
%%	\vspace{-2mm}
%\end{table}

%\vspace{-2mm}
\section{Conclusions}
%\vspace{-2mm}

%Though inspiring, big learning also shares the same constraints of foundation models \cite{bommasani2021opportunities,yuan2022roadmap}; \eg either to comprehensively analyze its properties or to verify its effectiveness (for general downstream tasks) is extremely challenging and time-consuming.
%Therefore, we believe that big learning needs our community and vice versa.

Cooperation plays a key role in the success of foundation models.
Based on this, big cooperative learning is proposed, which unifies the training of foundation models and delivers remarkable power of exploration.
Tailored $2$-D simulations and diverse experiments are conducted to demonstrate its principle, which in turn explicitly justifies the success of foundation models from the perspective of learning.
Big learning is a new dimension for upgrading conventional machine learning paradigms.

\begin{ack}
	This work was supported in part by the Key Basic Research Project of the Foundation Strengthening Program (2023-JCJQ-ZD-123-06), the Natural Science Foundation of Guangdong Province (2024A1515010221), and the Guangdong Provincial Pearl River Talents Program (2019ZT08X751).
%	Use unnumbered first level headings for the acknowledgments. All acknowledgments
%	go at the end of the paper before the list of references. Moreover, you are required to declare
%	funding (financial activities supporting the submitted work) and competing interests (related financial activities outside the submitted work).
%	More information about this disclosure can be found at: \url{https://neurips.cc/Conferences/2024/PaperInformation/FundingDisclosure}.
%	
%	
%	Do {\bf not} include this section in the anonymized submission, only in the final paper. You can use the \texttt{ack} environment provided in the style file to automatically hide this section in the anonymized submission.
\end{ack}

%\section*{References}
%
%References follow the acknowledgments in the camera-ready paper. Use unnumbered first-level heading for
%the references. Any choice of citation style is acceptable as long as you are
%consistent. It is permissible to reduce the font size to \verb+small+ (9 point)
%when listing the references.
%Note that the Reference section does not count towards the page limit.
%\medskip

%%%%%%%%%%%%%%%%%%%%%%%%%%%%%%%%%%%%%%%%%%%%%%%%%%%%%%%%%%%%
{\small
%	\bibliography{E:/Dropbox/ReferencesCong}
	\bibliography{ReferencesCong}

\begin{thebibliography}{10}

\bibitem{abdin2024phi}
Marah Abdin, Sam~Ade Jacobs, Ammar~Ahmad Awan, Jyoti Aneja, Ahmed Awadallah,
  Hany Awadalla, Nguyen Bach, Amit Bahree, Arash Bakhtiari, Harkirat Behl,
  et~al.
\newblock Phi-3 technical report: A highly capable language model locally on
  your phone.
\newblock {\em arXiv preprint arXiv:2404.14219}, 2024.

\bibitem{llama3modelcard}
AI@Meta.
\newblock Llama 3 model card.
\newblock 2024.

\bibitem{claude3}
Anthropic.
\newblock Introducing the next generation of claude, 2022.

\bibitem{bao2023one}
Fan Bao, Shen Nie, Kaiwen Xue, Chongxuan Li, Shi Pu, Yaole Wang, Gang Yue, Yue
  Cao, Hang Su, and Jun Zhu.
\newblock One transformer fits all distributions in multi-modal diffusion at
  scale.
\newblock {\em arXiv preprint arXiv:2303.06555}, 2023.

\bibitem{bao2021beit}
Hangbo Bao, Li~Dong, and Furu Wei.
\newblock Beit: Bert pre-training of image transformers.
\newblock {\em arXiv preprint arXiv:2106.08254}, 2021.

\bibitem{DallE3}
James Betker, Gabriel Goh, Li~Jing, Tim Brooks, Jianfeng Wang, Linjie Li, Long
  Ouyang, Juntang Zhuang, Joyce Lee, Yufei Guo, Wesam Manassra, Prafulla
  Dhariwal, Casey Chu, Yunxin Jiao, and Aditya Ramesh.
\newblock Improving image generation with better captions.
\newblock 2024.

\bibitem{bommasani2021opportunities}
Rishi Bommasani, Drew~A Hudson, Ehsan Adeli, Russ Altman, Simran Arora, Sydney
  von Arx, Michael~S Bernstein, Jeannette Bohg, Antoine Bosselut, Emma
  Brunskill, et~al.
\newblock On the opportunities and risks of foundation models.
\newblock {\em arXiv preprint arXiv:2108.07258}, 2021.

\bibitem{videoworldsimulators2024}
Tim Brooks, Bill Peebles, Connor Holmes, Will DePue, Yufei Guo, Li~Jing, David
  Schnurr, Joe Taylor, Troy Luhman, Eric Luhman, Clarence Ng, Ricky Wang, and
  Aditya Ramesh.
\newblock Video generation models as world simulators.
\newblock 2024.

\bibitem{brown2020language}
Tom Brown, Benjamin Mann, Nick Ryder, Melanie Subbiah, Jared~D Kaplan, Prafulla
  Dhariwal, Arvind Neelakantan, Pranav Shyam, Girish Sastry, Amanda Askell,
  et~al.
\newblock Language models are few-shot learners.
\newblock {\em NeurIPS}, 33:1877--1901, 2020.

\bibitem{cong2024big}
Yulai Cong and Sijia Li.
\newblock Big learning expectation maximization.
\newblock In {\em Proceedings of the AAAI Conference on Artificial
  Intelligence}, volume~38, pages 11669--11677, 2024.

\bibitem{devlin2018bert}
J.~Devlin, M.~Chang, K.~Lee, and K.~Toutanova.
\newblock {BERT}: Pre-training of deep bidirectional transformers for language
  understanding.
\newblock {\em arXiv preprint arXiv:1810.04805}, 2018.

\bibitem{goodfellow2014generative}
I.~Goodfellow, J.~Pouget-Abadie, M.~Mirza, B.~Xu, D.~Warde-Farley, S.~Ozair,
  A.~Courville, and Y.~Bengio.
\newblock Generative adversarial nets.
\newblock In {\em NeurIPS}, pages 2672--2680, 2014.

\bibitem{he2021masked}
Kaiming He, Xinlei Chen, Saining Xie, Yanghao Li, Piotr Doll{\'a}r, and Ross
  Girshick.
\newblock Masked autoencoders are scalable vision learners.
\newblock {\em arXiv preprint arXiv:2111.06377}, 2021.

\bibitem{ho2020denoising}
Jonathan Ho, Ajay Jain, and Pieter Abbeel.
\newblock Denoising diffusion probabilistic models.
\newblock {\em NeurIPS}, 33:6840--6851, 2020.

\bibitem{johnson1991joining}
David~W Johnson and Frank~P Johnson.
\newblock {\em Joining together: Group theory and group skills}.
\newblock Prentice-Hall, Inc, 1991.

\bibitem{johnson1987learning}
David~W Johnson and Roger~T Johnson.
\newblock {\em Learning together and alone: Cooperative, competitive, and
  individualistic learning}.
\newblock Prentice-Hall, Inc, 1987.

\bibitem{johnson1989cooperation}
David~W Johnson and Roger~T Johnson.
\newblock {\em Cooperation and competition: Theory and research}.
\newblock Interaction Book Company, 1989.

\bibitem{roger1994overview}
Roger~T Johnson and David~W Johnson.
\newblock An overview of cooperative learning.
\newblock {\em Creativity and collaborative learning}, 14(2):1--21, 1994.

\bibitem{joshi2020spanbert}
Mandar Joshi, Danqi Chen, Yinhan Liu, Daniel~S Weld, Luke Zettlemoyer, and Omer
  Levy.
\newblock Spanbert: Improving pre-training by representing and predicting
  spans.
\newblock {\em Transactions of the association for computational linguistics},
  8:64--77, 2020.

\bibitem{kim2021lipschitz}
Hyunjik Kim, George Papamakarios, and Andriy Mnih.
\newblock The lipschitz constant of self-attention.
\newblock In {\em International Conference on Machine Learning}, pages
  5562--5571. PMLR, 2021.

\bibitem{lan2019albert}
Zhenzhong Lan, Mingda Chen, Sebastian Goodman, Kevin Gimpel, Piyush Sharma, and
  Radu Soricut.
\newblock Albert: A lite bert for self-supervised learning of language
  representations.
\newblock {\em arXiv preprint arXiv:1909.11942}, 2019.

\bibitem{lee2021vitgan}
Kwonjoon Lee, Huiwen Chang, Lu~Jiang, Han Zhang, Zhuowen Tu, and Ce~Liu.
\newblock Vitgan: Training gans with vision transformers.
\newblock {\em arXiv preprint arXiv:2107.04589}, 2021.

\bibitem{lindsay1995mixture}
Bruce~G Lindsay.
\newblock Mixture models: theory, geometry, and applications.
\newblock Ims, 1995.

\bibitem{liu2019roberta}
Y.~Liu, M.~Ott, N.~Goyal, J.~Du, M.~Joshi, D.~Chen, O.~Levy, M.~Lewis,
  L.~Zettlemoyer, and V.~Stoyanov.
\newblock {RoBERTa}: A robustly optimized bert pretraining approach.
\newblock {\em arXiv preprint arXiv:1907.11692}, 2019.

\bibitem{loshchilov2017decoupled}
Ilya Loshchilov and Frank Hutter.
\newblock Decoupled weight decay regularization.
\newblock {\em arXiv preprint arXiv:1711.05101}, 2017.

\bibitem{mescheder2018training}
L.~Mescheder, A.~Geiger, and S.~Nowozin.
\newblock Which training methods for {GANs} do actually converge?
\newblock In {\em ICML}, pages 3478--3487, 2018.

\bibitem{ChatGPT}
OpenAI.
\newblock Chatgpt: Optimizing language models for dialogue.
\newblock \url{https://openai.com/blog/chatgpt}, 2022.
\newblock Accessed: 2022-11-30.

\bibitem{GPT4}
OpenAI.
\newblock Gpt-4.
\newblock \url{https://openai.com/research/gpt-4}, 2023.
\newblock Accessed: 2023-03-14.

\bibitem{ouyang2022training}
Long Ouyang, Jeff Wu, Xu~Jiang, Diogo Almeida, Carroll~L Wainwright, Pamela
  Mishkin, Chong Zhang, Sandhini Agarwal, Katarina Slama, Alex Ray, et~al.
\newblock Training language models to follow instructions with human feedback.
\newblock {\em arXiv preprint arXiv:2203.02155}, 2022.

\bibitem{peel2000finite}
DAVID Peel and G~MacLahlan.
\newblock Finite mixture models.
\newblock {\em John \& Sons}, 2000.

\bibitem{radford2019language}
A.~Radford, J.~Wu, R.~Child, D.~Luan, D.~Amodei, and I.~Sutskever.
\newblock Language models are unsupervised multitask learners.
\newblock {\em OpenAI Blog}, 1(8):9, 2019.

\bibitem{ramesh2022hierarchical}
Aditya Ramesh, Prafulla Dhariwal, Alex Nichol, Casey Chu, and Mark Chen.
\newblock Hierarchical text-conditional image generation with clip latents.
\newblock {\em arXiv preprint arXiv:2204.06125}, 2022.

\bibitem{ramesh2021zero}
Aditya Ramesh, Mikhail Pavlov, Gabriel Goh, Scott Gray, Chelsea Voss, Alec
  Radford, Mark Chen, and Ilya Sutskever.
\newblock Zero-shot text-to-image generation.
\newblock In {\em ICML}, pages 8821--8831. PMLR, 2021.

\bibitem{reid2024gemini}
Machel Reid, Nikolay Savinov, Denis Teplyashin, Dmitry Lepikhin, Timothy
  Lillicrap, Jean-baptiste Alayrac, Radu Soricut, Angeliki Lazaridou, Orhan
  Firat, Julian Schrittwieser, et~al.
\newblock Gemini 1.5: Unlocking multimodal understanding across millions of
  tokens of context.
\newblock {\em arXiv preprint arXiv:2403.05530}, 2024.

\bibitem{rombach2022high}
Robin Rombach, Andreas Blattmann, Dominik Lorenz, Patrick Esser, and Bj{\"o}rn
  Ommer.
\newblock High-resolution image synthesis with latent diffusion models.
\newblock In {\em CVPR}, pages 10684--10695, 2022.

\bibitem{slavin1980cooperative}
Robert~E Slavin.
\newblock Cooperative learning.
\newblock {\em Review of educational research}, 50(2):315--342, 1980.

\bibitem{song2020score}
Yang Song, Jascha Sohl-Dickstein, Diederik~P Kingma, Abhishek Kumar, Stefano
  Ermon, and Ben Poole.
\newblock Score-based generative modeling through stochastic differential
  equations.
\newblock {\em arXiv preprint arXiv:2011.13456}, 2020.

\bibitem{stickland2019bert}
A.~Stickland and I.~Murray.
\newblock {BERT} and {PALs}: Projected attention layers for efficient
  adaptation in multi-task learning.
\newblock {\em arXiv preprint arXiv:1902.02671}, 2019.

\bibitem{tamkin2021dabs}
Alex Tamkin, Vincent Liu, Rongfei Lu, Daniel Fein, Colin Schultz, and Noah
  Goodman.
\newblock {DABS}: A domain-agnostic benchmark for self-supervised learning.
\newblock {\em arXiv preprint arXiv:2111.12062}, 2021.

\bibitem{team2023gemini}
Gemini Team, Rohan Anil, Sebastian Borgeaud, Yonghui Wu, Jean-Baptiste Alayrac,
  Jiahui Yu, Radu Soricut, Johan Schalkwyk, Andrew~M Dai, Anja Hauth, et~al.
\newblock Gemini: a family of highly capable multimodal models.
\newblock {\em arXiv preprint arXiv:2312.11805}, 2023.

\bibitem{tian2024visual}
Keyu Tian, Yi~Jiang, Zehuan Yuan, Bingyue Peng, and Liwei Wang.
\newblock Visual autoregressive modeling: Scalable image generation via
  next-scale prediction.
\newblock {\em arXiv preprint arXiv:2404.02905}, 2024.

\bibitem{touvron2021going}
Hugo Touvron, Matthieu Cord, Alexandre Sablayrolles, Gabriel Synnaeve, and
  Herv{\'e} J{\'e}gou.
\newblock Going deeper with image transformers.
\newblock In {\em ICCV}, pages 32--42, 2021.

\bibitem{vaswani2017attention}
A.~Vaswani, N.~Shazeer, N.~Parmar, J.~Uszkoreit, L.~Jones, A.~Gomez,
  {\L}.~Kaiser, and I.~Polosukhin.
\newblock Attention is all you need.
\newblock In {\em NeurIPS}, pages 5998--6008, 2017.

\bibitem{wang2018glue}
A.~Wang, A.~Singh, J.~Michael, F.~Hill, O.~Levy, and S.~Bowman.
\newblock Glue: A multi-task benchmark and analysis platform for natural
  language understanding.
\newblock In {\em ICLR}, 2018.

\bibitem{wei2021masked}
Chen Wei, Haoqi Fan, Saining Xie, Chao-Yuan Wu, Alan Yuille, and Christoph
  Feichtenhofer.
\newblock Masked feature prediction for self-supervised visual pre-training.
\newblock {\em arXiv preprint arXiv:2112.09133}, 2021.

\bibitem{wolf-etal-2020-transformers}
Thomas Wolf, Lysandre Debut, Victor Sanh, Julien Chaumond, Clement Delangue,
  Anthony Moi, Pierric Cistac, Tim Rault, Rémi Louf, Morgan Funtowicz, Joe
  Davison, Sam Shleifer, Patrick von Platen, Clara Ma, Yacine Jernite, Julien
  Plu, Canwen Xu, Teven~Le Scao, Sylvain Gugger, Mariama Drame, Quentin Lhoest,
  and Alexander~M. Rush.
\newblock Transformers: State-of-the-art natural language processing.
\newblock In {\em Proceedings of the 2020 Conference on Empirical Methods in
  Natural Language Processing: System Demonstrations}, pages 38--45, Online,
  October 2020. Association for Computational Linguistics.

\bibitem{yang2019xlnet}
Z.~Yang, Z.~Dai, Y.~Yang, J.~Carbonell, R.~Salakhutdinov, and Q.~Le.
\newblock {XLNet}: Generalized autoregressive pretraining for language
  understanding.
\newblock In {\em NeurIPS}, pages 5753--5763, 2019.

\bibitem{yuan2022roadmap}
Sha Yuan, Hanyu Zhao, Shuai Zhao, Jiahong Leng, Yangxiao Liang, Xiaozhi Wang,
  Jifan Yu, Xin Lv, Zhou Shao, Jiaao He, et~al.
\newblock A roadmap for big model.
\newblock {\em arXiv preprint arXiv:2203.14101}, 2022.

\end{thebibliography}
	%	\bibliography{ReferencesCong}
	%	\bibliography{D:/Dropbox/[cvpr2020]GAN_fix/ReferencesCong}
	\bibliographystyle{plain}
}

\newpage
\appendix

\begin{figure}[H]
%	\vspace{-5mm}
	\centering
	\subfloat[A data can be exploited from many perspectives.
		When given a complete/incomplete data sample $\xv \sim q(\xv)$, one simultaneously receives multiple joint, marginal, and conditional samples from $q(\xv_{\Tbb}|\xv_{\Sbb}), \forall (\Sbb,\Tbb)$.
		These samples contain valuable data information associated with \eg data manifold and correlation among data patches (or words in text applications). Since they all demonstrate the \emph{unique} underlying data distribution $q(\xv)$ (despite from diverse different perspectives), there is room with potential for introducing implicit regularizations among them via consistent multi-task training, \ie the presented big cooperative learning.
	]{
		\includegraphics[width=0.9\columnwidth]{Figures/data_multi_perspective.png}
		\label{fig:}}
	\\
	\subfloat[
		Conventional machine learning, \ie joint learning with only the complete data, cannot fully exploit the data information, \eg the diverse correlations among data patches within conditional data and those within the incomplete data samples. 
		Accordingly, only single joint data capability can be learned by the model. 
	]{
		\includegraphics[height=0.33\columnwidth]{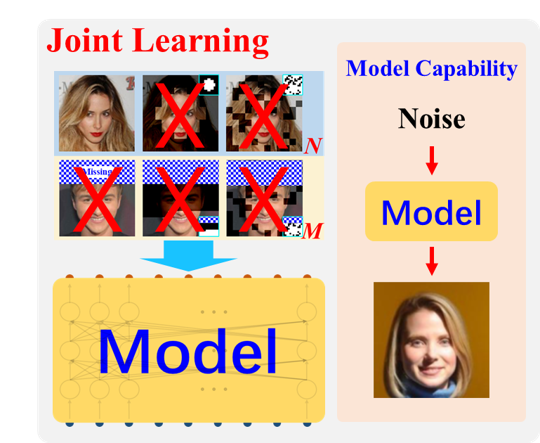}
		\label{fig:}}
	\qquad
	\subfloat[
		The big learning flexibly and comprehensively exploits the diverse joint, marginal, and conditional samples inherent in complete and incomplete training data, leading to a consistent, unified, and principled learning framework underlying most foundation models.
		Besides, the big learning naturally delivers many/all joint, marginal, and conditional data capabilities across potentially diverse domains with a universal model.
	]{
		\includegraphics[height=0.33\columnwidth]{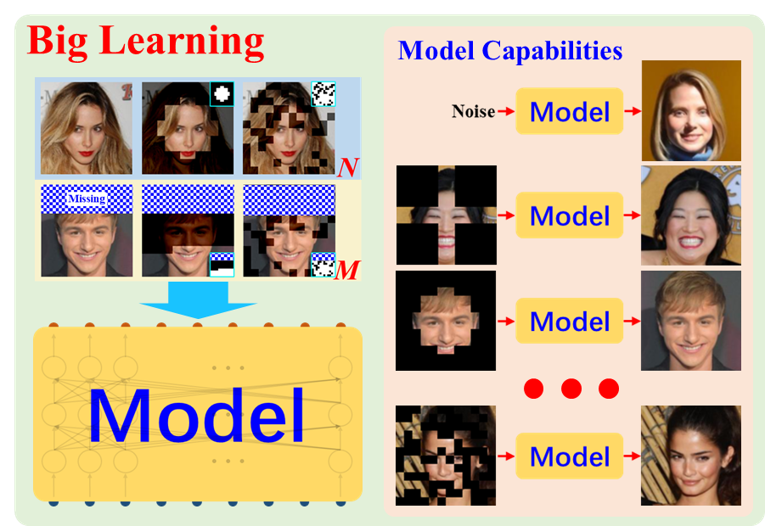}
		\label{fig:}}
	\caption{Big picture of big cooperative learning. }
	\label{appfig:bigpicture}
%	\vspace{-2mm}
\end{figure}

\section{Big Cooperative Learning With Multi-Modal Data}
\label{appsec:big_learning_multimodal}

For simplicity, we first consider multi-modal applications where all the modals of data $\xv=[\xv',\yv',\zv',\cdots]$ have the same data type (like images). 
In such situations, one may naively conduct big cooperative learning in the $\xv$-space or its transformed $\Xv$-space, as discussed in the main manuscript.

However, for general multi-modal applications, each modal often has a different data type; for example, $\xv'$ is a continuous image while $\yv'$ represents discrete texts.
In these situations, one may consider 
$(i)$ employing modal-specific transformations to align all the modals in a latent space followed by big cooperative learning in that latent space mimicking \cite{ramesh2021zero,rombach2022high} or 
$(ii)$ to implicitly model $p_{\thetav}(\xv_{\Tbb}|\xv_{\Sbb})$ (exampled with $\Xv=\xv$) via decomposed components, \eg 
\beq\bali\label{eq:biglearn_doubleFP}
p_{\thetav}(\xv_{\Tbb}|\xv_{\Sbb}) = 
p_{\thetav}(\xv'_{\Tbb_{1}}|\xv_{\Sbb})
p_{\thetav}(\yv'_{\Tbb_{2}}|\xv'_{\Tbb_{1}},\xv_{\Sbb})
p_{\thetav}(\zv'_{\Tbb_{3}}|\xv'_{\Tbb_{1}},\yv'_{\Tbb_{2}},\xv_{\Sbb}),
\eali\eeq
where each decomposed component has one unique data type, $\Tbb_{1}/\Tbb_{2}/\Tbb_{3}$ denotes the target indexes of each modal $\xv'/\yv'/\zv'$,  $\Tbb=[\Tbb_{1},\Tbb_{2},\Tbb_{3}]$, the modeling of each modal (\eg $p_{\thetav}(\yv'_{\Tbb_{2}}|\xv'_{\Tbb_{1}},\xv_{\Sbb})$) may be different, and one may separately train each decomposed component with modal-specific divergence/distance (\ie the arrow ``$\rightarrow$''). Note to maintain cooperations among the training associated with each modal is essential for a successful big cooperative learning.

A significant body of research remains to be conducted.

\section{Details and Interesting Side-Products of Tailored $2$-D Simulations}
\label{appsec:tailored_2D_simulation}

With a parameterized model $p_{\thetav}(\xv)$ to approximate the underlying (data) distribution $q(\xv)$, two popular learning paradigms are represented by the mode-covering forward KL minimization $\min_{\thetav} \KL[q(\xv)||p_{\thetav}(\xv)]$ (\eg maximum log-likelihood learning) and the mode-seeking reverse KL minimization $\min_{\thetav} \KL[p_{\thetav}(\xv)||q(\xv)]$ (closely related to adversarial learning).

As discussed in the main manuscript, we demonstrate the principle of big cooperative learning via the mode-seeking reverse KL minimization in tailored situations with $2$-GMMs. 
The simulation (see Section \ref{sec:2D_gmm_simulation} of the main manuscript) is actually an optimization issue with $1$-dimensional observation (\ie only the first dimension of $\xv$ is associated with the parameters $\mu_1$ and $\mu_2$ that constitute $\thetav$) put in a higher $2$-dimensional observation space ($\xv\in \Rbb^2$).

All the analyses are possible because of the simplicity of GMMs, \ie $(i)$ GMMs have analytical marginal, conditional, transformed marginal, and transformed conditional distributions, and $(ii)$ the result of a GMM convolved with a Gaussian distribution (\ie noising) is also analytically expressed.
On the other hand, GMMs with sufficient components are also powerful enough to approximate \emph{any} continuous distribution arbitrarily well \cite{lindsay1995mixture,peel2000finite}, justifying the representativeness of the tailored simulations in Sections \ref{sec:2D_gmm_simulation} and \ref{sec:25GMM_exploration} of the main manuscript.

\begin{figure}[H]
	%	\vspace{-5mm}
	\centering
	\subfloat[$q(\xv)$]{
		\includegraphics[height=0.3\columnwidth]{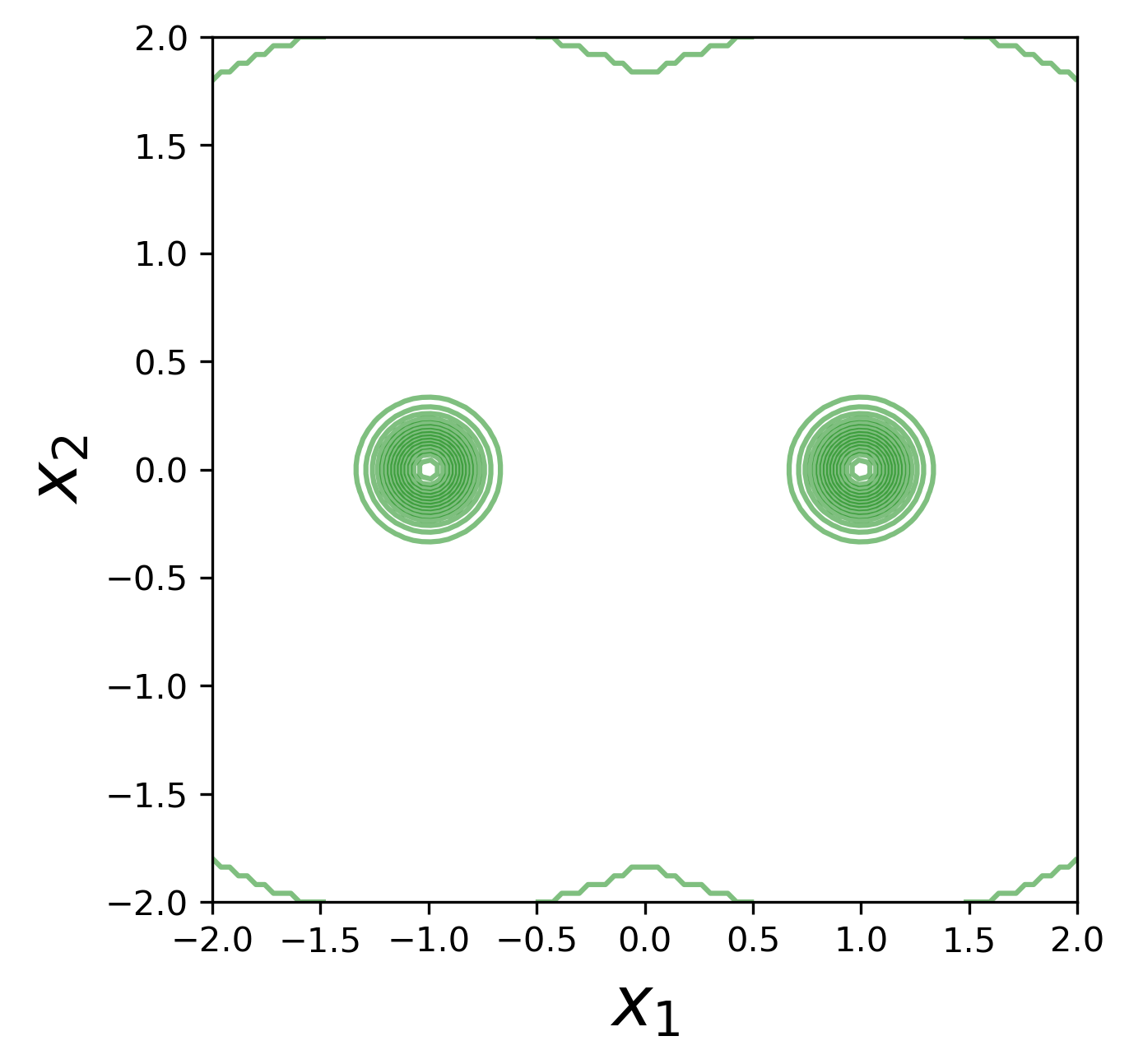}
		\label{fig:}}
	\qquad\qquad
	\subfloat[Joint Reverse $\KL(p_{\thetav}(\xv)||q(\xv))$]{
		\includegraphics[height=0.3\columnwidth]{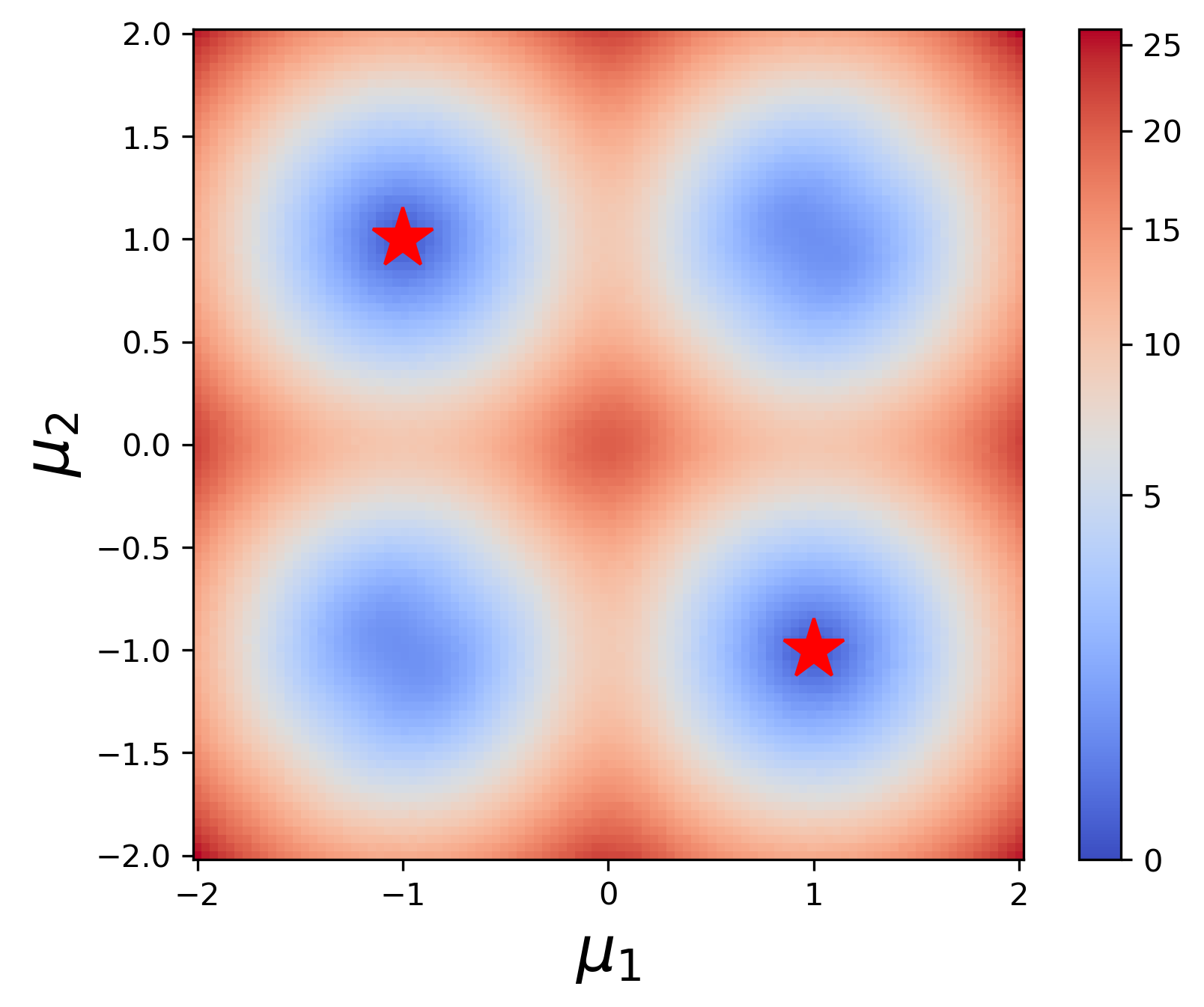}
		\label{fig:}}
	\caption{Demonstrations of the underlying distribution $q(\xv)$ (a) and the naive joint reverse KL objective $\KL[p_{\thetav}(\xv)||q(\xv)]$ (b). $\sigma^2=0.02$.
		The two global optima are marked with red stars in (b), where two local optima are also prominent.}
	\label{fig:qx_jointKL}
	%	\vspace{-2mm}
\end{figure}

Fig. \ref{fig:qx_jointKL} shows the objective surface of joint matching with $\KL[p_{\thetav}(\xv)||q(\xv)]$, where $\sigma^2=0.02$ and significant local optima emerge. 
In fact, the significance of local optima increases with the decreasing of $\sigma$ (see the difference between Fig. \ref{fig:qx_jointKL} and Fig. \ref{fig:naive_JMC_not_work} of the main manuscript).

\begin{figure}[H]
	\centering
	\subfloat[$\KL(p_{\thetav}(y_1)||q(y_1)),\yv=\Amat\xv$ with diverse rotation $\Amat$]{
		\includegraphics[height=0.38\columnwidth]{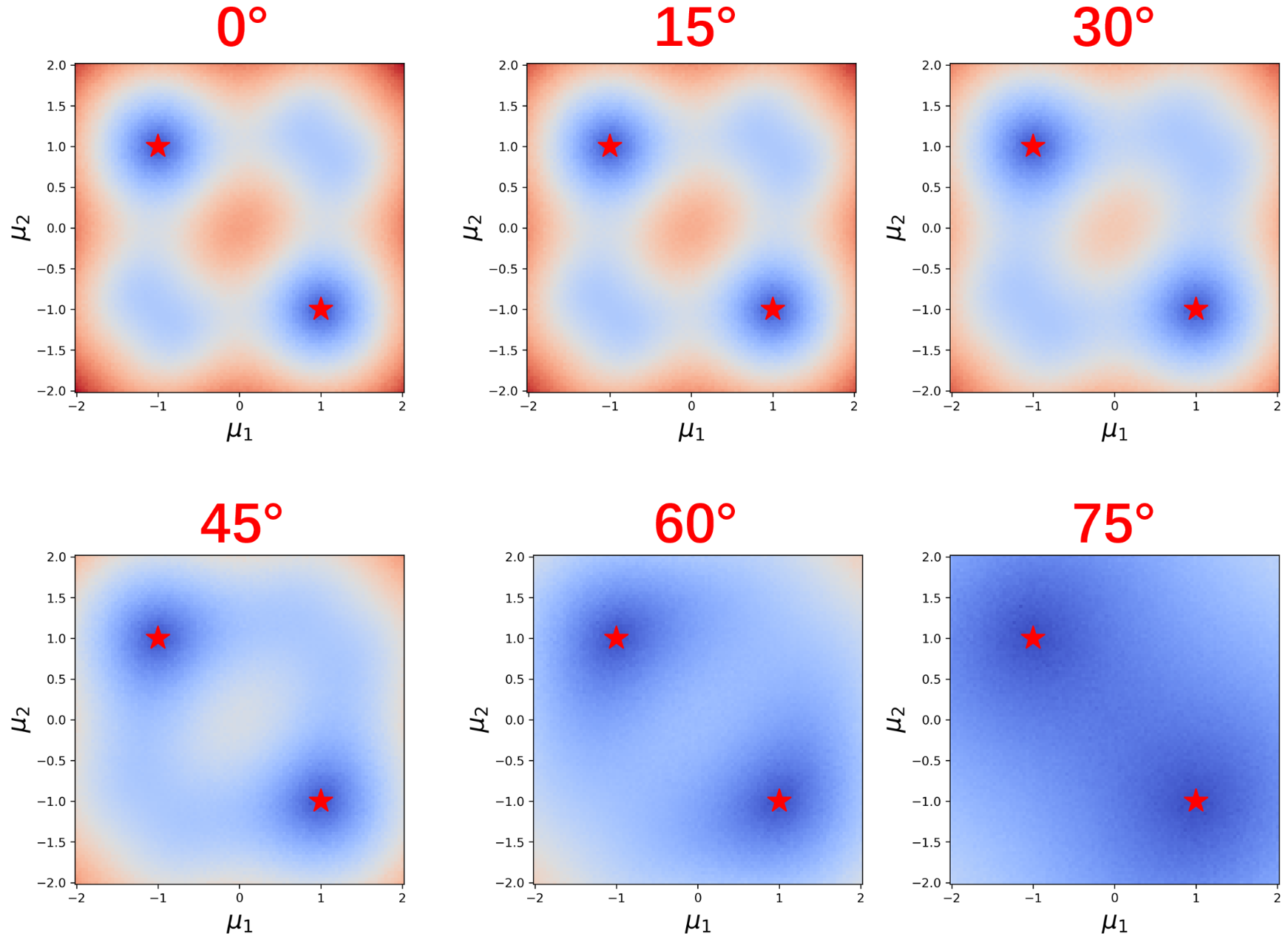}
		\label{appfig:margin_match_diverse_rotation}}
	\quad
	\subfloat[$\Ebb_{U(\Amat)}\KL(p_{\thetav}(y_1)||q(y_1))$]{
		\includegraphics[height=0.38\columnwidth]{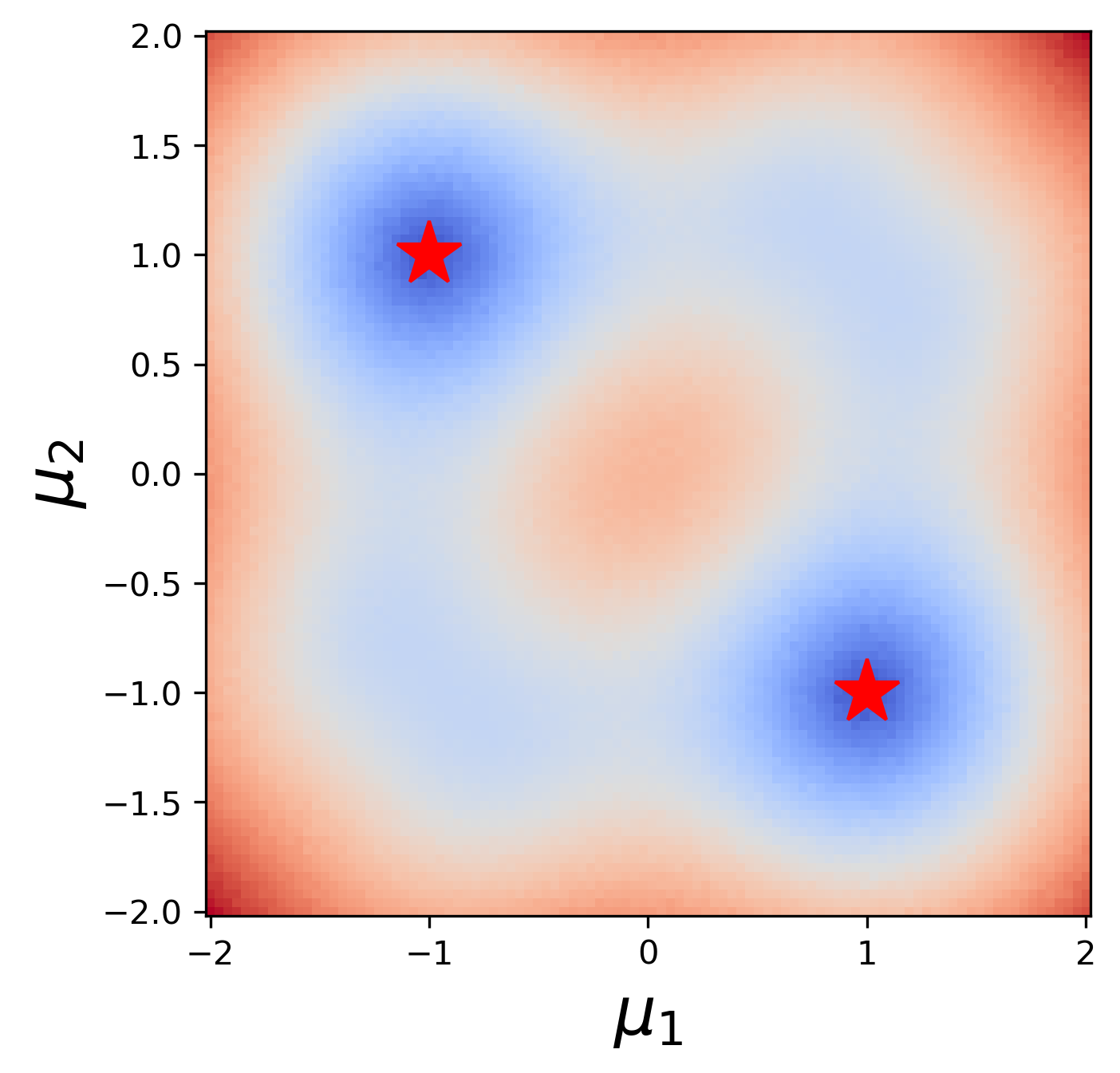}
		\label{appfig:averaged_margin_match}}
	\caption{Demonstration of the preference of a bi-level optimization when using transformed marginal matchings.
		(a) Transformed marginal matching has an magnitude correlated with the significance of local optima.
		(b) Optimization with an uniformly sampled marginal matching may still suffer from local optima, where a bi-level optimization would be beneficial. 
		See the supplementary ``Figure\_video\_margin.gif.''
	}
	\label{appfig:side_product_margin}
\end{figure}

\begin{figure}[H]
	\centering
	\subfloat[$\KL(p_{\thetav}(y_1|y_2)||q(y_1|y_2)),\yv=\Amat\xv$ with diverse rotation $\Amat$ and different settings of $y_2$]{
		\includegraphics[height=0.42\columnwidth]{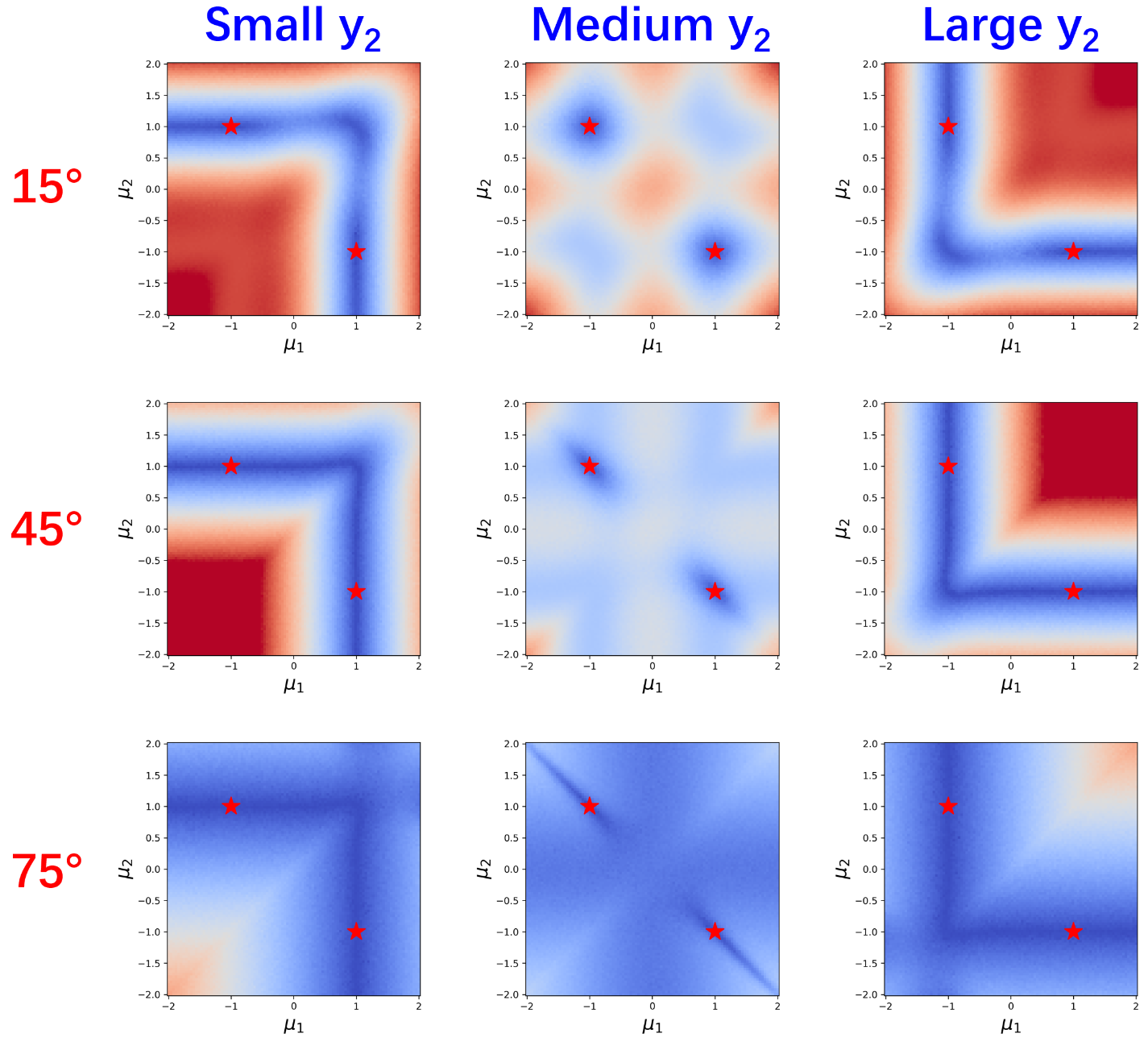}
		\label{appfig:condition_match_diverse_rotation}}
	\qquad
	\subfloat[$\Ebb_{U(\Amat)U(y_2)}\KL(p_{\thetav}(y_1|y_2)||q(y_1|y_2))$]{
		\includegraphics[height=0.42\columnwidth]{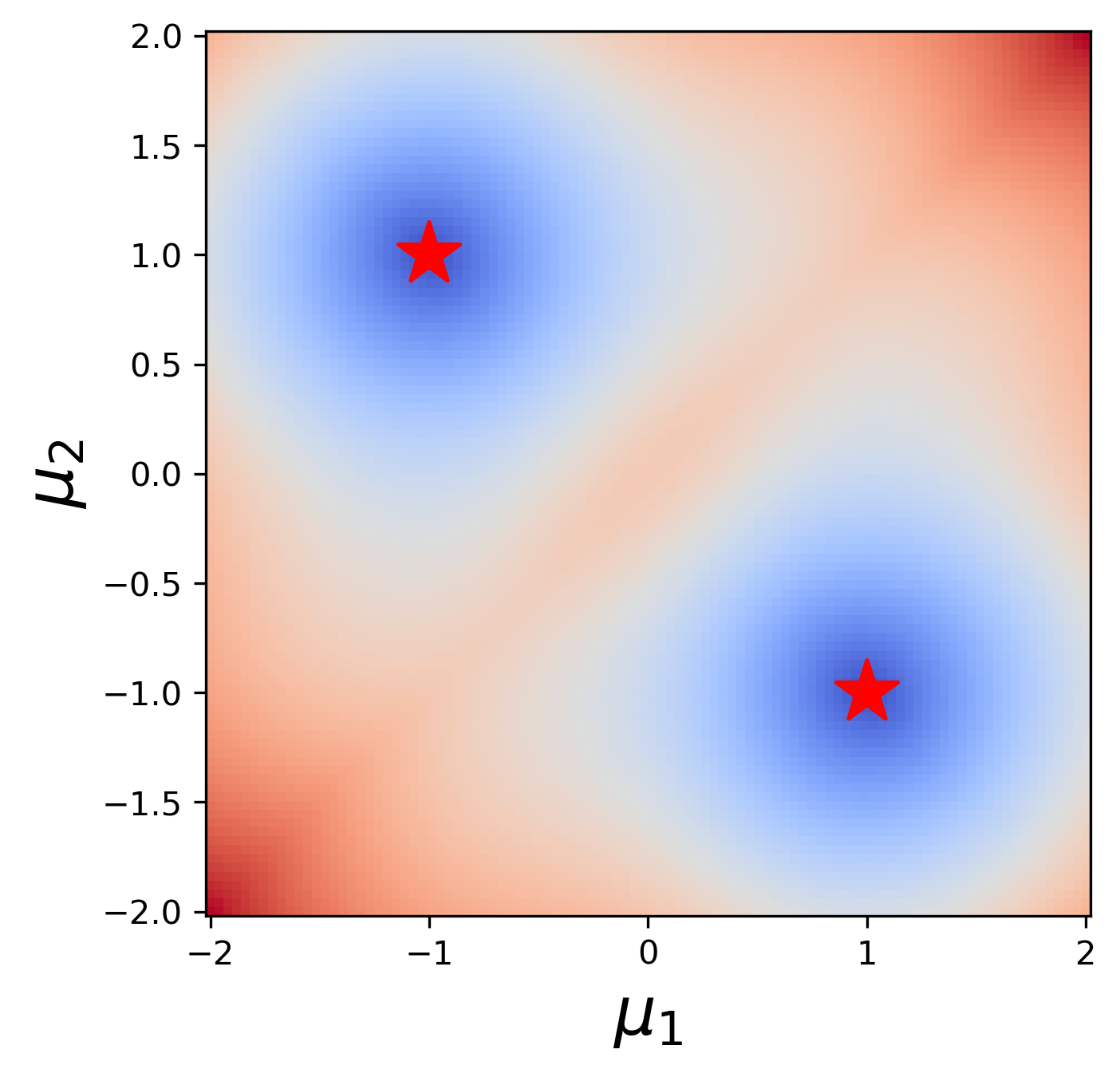}
		\label{appfig:averaged_condition_match}}
	\caption{Demonstration of the principle of diverse conditional matchings.
		(a) Transformed conditional matching has a loss surface diversely changing with different rotations and conditions.
		(b) Optimization with an uniformly sampled conditional matching may overlook local optima, delivering an appealing averaged loss surface. 
		See the supplementary ``Figure\_video\_condition.gif.''
	}
	\label{appfig:side_product_condition}
\end{figure}

\begin{figure}[H]
	%	\vspace{-5mm}
	\centering
	\includegraphics[width=\columnwidth]{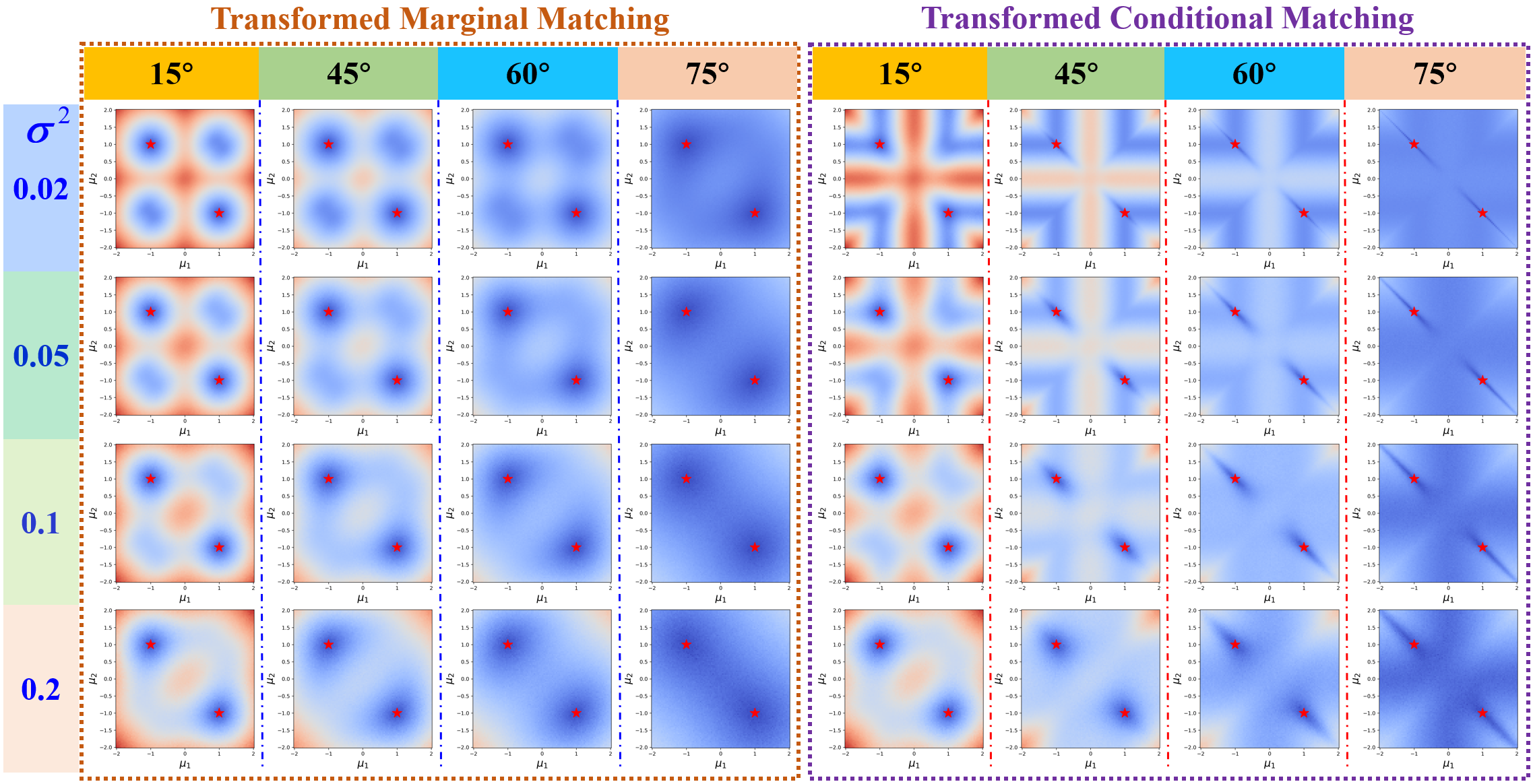}
	\caption{Demonstration of the power of multi-scale noising.
		Multi-scale noising (when applicable) serves as a new dimension for data-level transformations of big cooperative learning.
		Note the local optima gradually vanish with the increasing noise variance $\sigma^2$, but the local surfaces surrounding the global optima are also gradually flattened.
		Different characteristics among multi-scale noising enable cooperation.
	}
	\label{appfig:side_product_multinoise}
	%	\vspace{-2mm}
\end{figure}

In addition to $(i)$ Fig. \ref{fig:naive_JMC_not_work} showing that joint, marginal, and conditional matchings have no advantages over joint matching for situations with independent features/tokens and $(ii)$ Fig. \ref{fig:trans_MC_matching} showing dependency among tokens (introduced via rotational transformations) enables cooperation among diverse matchings (local optima unstably vary but global optima are stably consistent), we reveal below several interesting side-products that potentially benefit implementations of improved foundation models.
\begin{itemize}%[leftmargin=5mm]	
	\item Fig. \ref{appfig:side_product_margin} demonstrates that a bi-level optimization may be preferred when using transformed marginal (and joint) matchings to conduct big cooperative learning.
	
	\item Fig. \ref{appfig:side_product_condition} shows that averaging over diverse transformed conditional matchings may overlook local optima and a single-level optimization is sufficient.
	
	\item Fig. \ref{appfig:side_product_multinoise} illustrates that multi-scale noising (when applicable) serves as powerful data-level transformations to enable cooperation for big cooperative learning.
	
\end{itemize}

\section{More Details of the $25$-GMM Reverse-KL-Minimization Simulation}
\label{appsec:25GMM_exploration}

\begin{figure}[H]
	\centering
	\includegraphics[height=0.42\columnwidth]{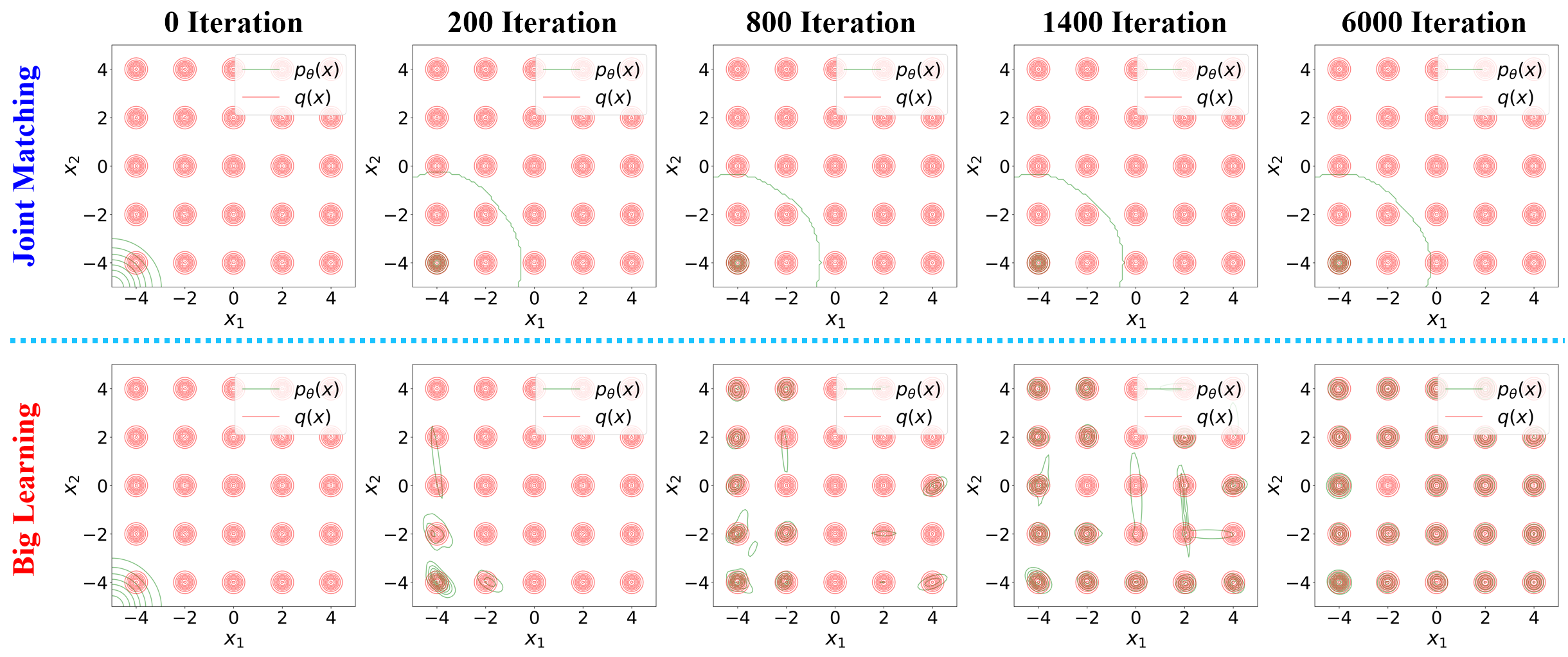}
	\caption{Demonstration of the amazing exploration power of big cooperative learning.
		See the supplementary ``Figure\_video\_25GMM\_BLexloration.gif''.
	}
	\label{appfig:25GMM_Exploration_app}
\end{figure}

Similar to the tailored $2$-D simulation of Section \ref{sec:2D_gmm_simulation} of the main manuscript, we set both $q(\xv)$ and $p_{\thetav}(\xv)$ as a GMM with $25$ components (\ie a $25$-GMM) and concern minimizing the differences between them via reverse-KL minimization. 
Specifically, 
\beq
q(\xv) = p_{\thetav^{*}}(\xv) = \sum\nolimits_{i=1}^{25} \frac{1}{25} \Nc(\xv|\muv^{*}_i, \Sigmamat^{*}_i)
\quad \text{and} \quad
p_{\thetav}(\xv) = \sum\nolimits_{i=1}^{25} \frac{1}{25} \Nc(\xv|\muv_i, \Sigmamat_i),
\eeq
where $\muv^{*}$ is set as shown in Fig. \ref{appfig:25GMM_Exploration_app},  $\Sigmamat^{*}=\sigma^2 \Imat$ with $\sigma^2=0.05$, $\Sigmamat_i = \Lmat_i \Lmat_i^T$ with trainable lower-triangular $\Lmat_i$, and $\thetav=[\{\mu_i\}_{i=1}^{25},\{\Lmat_i\}_{i=1}^{25}]^T$.
$\{\mu_i\}$s are randomly initialized with $\Nc(-5, 0.01)$.

For training, we employ a SGD optimizer with a learning rate of $0.1$; we use $100$ samples for diverse reverse-KL minimization (\eg $100$ $\xv$s from $p_{\thetav}(\xv)$ for joint reverse-KL matching); we also adopt random orthogonal transformations (\ie $\yv=\Bmat\xv$ with $\Bmat$ denoting an orthogonal transformation) for transformed marginal and conditional matchings.

We only employ diverse joint and transformed marginal matchings to conduct big cooperative learning, because we empirically find it sufficient. 
We separate our big cooperative learning into two phases. In the first burn-in phase, we employ transformed marginal matchings to maximize exploration, and after that, we use both joint and transformed marginal matchings in the second phase, with probabilities of $[0.1, 0.9]$, respectively.

Fig. \ref{appfig:25GMM_Exploration_app} shows the results from naive joint reverse-KL matching and our big cooperative learning. 
It's evident that cooperation among diverse matchings delivers amazing exploration power that overlooks local optima, even each matching task is a mode-seeking reverse-KL minimization that often suffers from mode collapse.

\section{BigLearn-GAN: Big Cooperative Learning Generative Adversarial Nets}
\label{appsec:BigLearn_GAN}

\subsection{Network Architectures of BigLearn-GAN}
\label{appsec:BigLearn_GAN_archtecture}

Below we discuss the employed network architectures for the BigLearn-GAN generator and discriminator in Eq. \eqref{eq:model_to_data_all} of the main manuscript.

\begin{figure}[H]
	\centering
	\subfloat[BigLearn-GAN Generator]{
		\includegraphics[height=0.45\columnwidth]{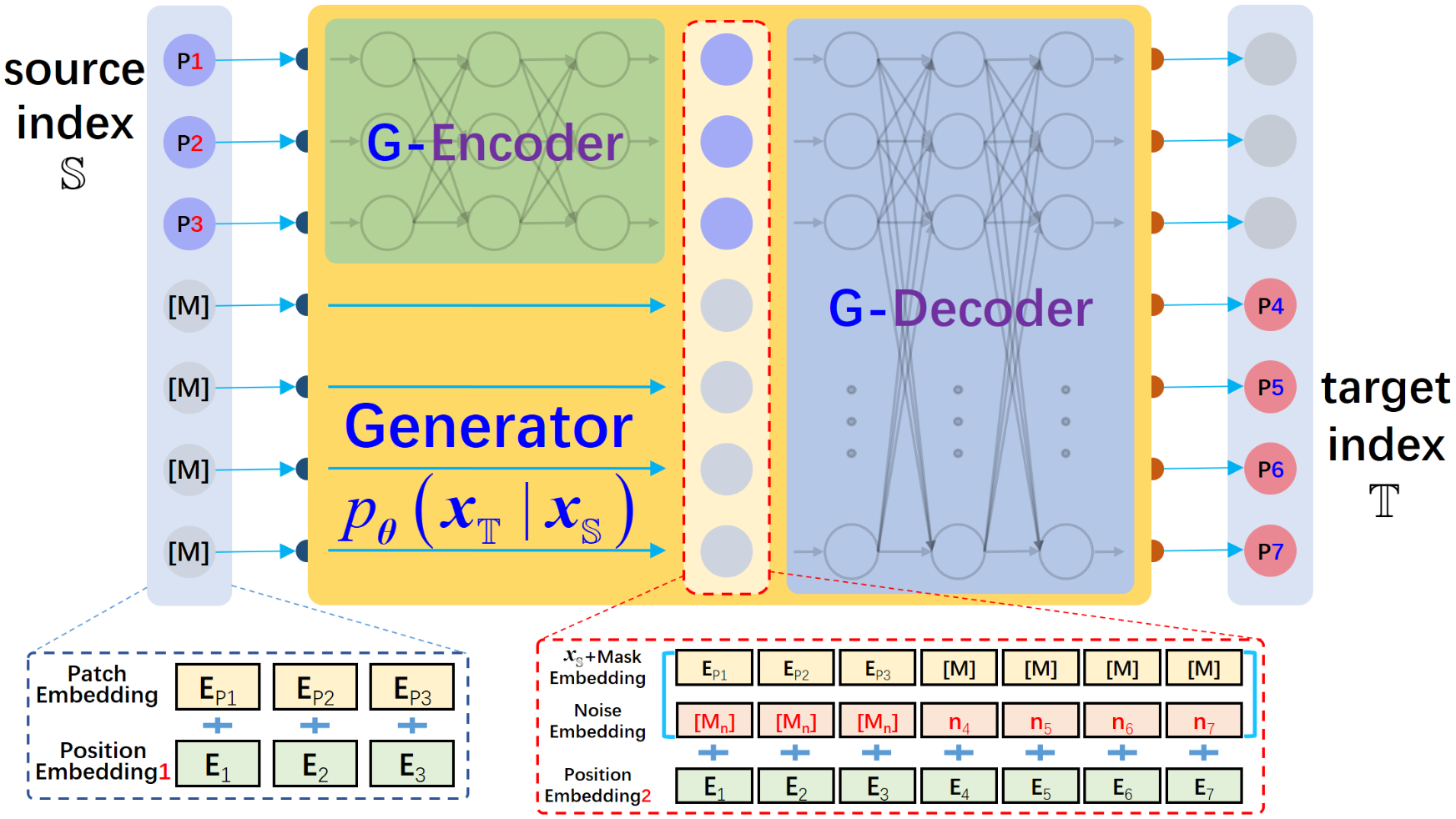}
		\label{appfig:arch_GANgenerator}}
	\qquad
	\subfloat[BigLearn-GAN Discriminator]{
		\includegraphics[height=0.45\columnwidth]{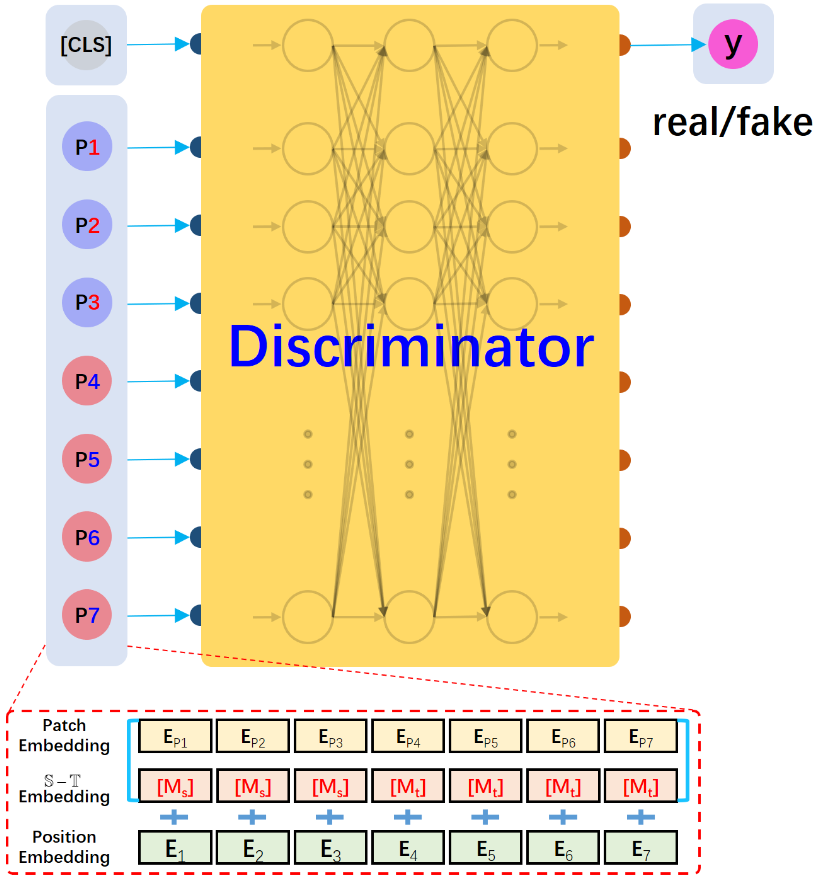}
		\label{appfig:arch_GANdiscriminiator}}
	\caption{Demonstrations of the implemented BigLearn-GAN generator and discriminator. 
	}
	\label{appfig:arch_GAN}
\end{figure}

Fig. \ref{appfig:arch_GAN} demonstrates the employed BigLearn-GAN generator and discriminator, both of which are constructed with Transformers to exploit their modeling capabilities and flexibilities.
\begin{itemize}
	\item \textbf{BigLearn-GAN Generator.} 
	Following the MAE \cite{he2021masked}, we design the universal generator $p_{\thetav}(\xv_{\Tbb} | \xv_{\Sbb})$ with an autoencoder-like architecture, which employs an encoding G-Encoder and a decoding G-Decoder, as shown in Fig. \ref{appfig:arch_GANgenerator}. 
	The G-Encoder encodes the source patches $\xv_{\Sbb}$ (if any) to their latent codes; then, these codes are combined with the mask tokens \texttt{[M]}, patch-wise noise embeddings, and new positional encodings to serve as the input of the G-Decoder; finally, the G-Decoder transforms its input to generate the target patches $\xv_{\Tbb}$. 
	
	$\texttt{[M]}$ tokens are inserted later in a middle layer, because doing this often improves performance and lowers the computational burden \cite{touvron2021going,he2021masked}.
	A noise $\zv$ is mapped with an $8$-layer MLP to produce the patch-wise noise embeddings $\{\nv_1,\cdots,\nv_L\}$. 
	Note we also introduce another toke $\texttt{[M}_{\texttt{n}}{]}$ to indicate no noise embeddings are necessary at the corresponding source locations in $\Sbb$.
	
	\item \textbf{BigLearn-GAN Discriminator.} 
	As shown in Fig. \ref{appfig:arch_GANdiscriminiator}, we also modify the Transformer architecture to construct the universal discriminator $\sigma(f_{\phiv}(\xv;\Sbb,\Tbb))$ that applies to all $(\Sbb,\Tbb)$ cases. 
	We employ an additional \texttt{CLS} token mimicking the BERT, whose output indicates whether the input patches are realistic or not (more specifically, whether they form a ``real'' data from $q(\xv_{\Tbb} | \xv_{\Sbb})$ or a fake one from $p_{\thetav}(\xv_{\Tbb} | \xv_{\Sbb})$, by referring to Eq. \eqref{eq:model_to_data_all} of the main manuscript). 
	The input of the discriminator consists of patch embeddings, positional embeddings, and two new special tokens ($\texttt{[M}_{\texttt{s}}{]}$ and $\texttt{[M}_{\texttt{t}}{]}$) that indicate source or target patches mimicking the sentence tokens in the BERT.
	
\end{itemize}

\subsection{Additional ``Communications'' in Big Cooperative Learning}
\label{appsec:BigLearn_GAN_communicate}

We notice that Eq. \ref{eq:model_to_data_all} of the main manuscript is equivalent to 
\beq\label{eq:model_to_data_all_app}
\min_{\thetav} \max_{\phiv} 
\Ebb_{q(\Sbb,\Tbb)}\big[
\Ebb_{q(\xv_{\Sbb\cup\Tbb})} {\log} \sigma[f_{\phiv}(\xv;\Sbb,\Tbb)] + 
\Ebb_{p_{\thetav}(\xv_{\Tbb} | \xv_{\Sbb}) q(\xv_{\Sbb})} {\log} \sigma[-f_{\phiv}(\xv;\Sbb,\Tbb)]
\big],
\eeq
where the optimal $f_{\phi^{*}}(\xv;\Sbb,\Tbb) = {\log} \frac{q(\xv_{\Tbb} | \xv_{\Sbb})}{p_{\thetav}(\xv_{\Tbb} | \xv_{\Sbb})} = {\log} \frac{q(\xv_{\Sbb\cup\Tbb})}{p_{\thetav}(\xv_{\Tbb} | \xv_{\Sbb}) q(\xv_{\Sbb})}$.
Accordingly, Eq. \ref{eq:model_to_data_all} and Eq. \ref{eq:model_to_data_all_app} ideally  performs $\min_{\thetav} \max_{\phiv} \Ebb_{q(\Sbb,\Tbb)} \JS[q(\xv_{\Sbb\cup\Tbb})||p_{\thetav}(\xv_{\Tbb} | \xv_{\Sbb}) q(\xv_{\Sbb})]$, \ie encouraging the joint matchings between $q(\xv_{\Sbb\cup\Tbb})$ and $p_{\thetav}(\xv_{\Tbb} | \xv_{\Sbb}) q(\xv_{\Sbb})$ for different $(\Sbb,\Tbb)$s.

Noticing that the universal $p_{\thetav}(\xv_{\Tbb}|\xv_{\Sbb})$ possesses versatile data sampling capabilities, one may further expand the learning-task scope of big cooperative learning with ``communications'' (\ie matchings) between two model distributions $p_{\thetav}(\xv_{\Tbb_1} | \xv_{\Sbb_1}) q(\xv_{\Sbb_1})$ and $p_{\thetav}(\xv_{\Tbb_2} | \xv_{\Sbb_2}) q(\xv_{\Sbb_2})$ with $\Sbb^{1}\cup\Tbb^{1} = \Sbb^{2}\cup\Tbb^{2}$, because they share the same ultimate goal of matching $q(\xv_{\Sbb^{1}\cup\Tbb^{1}})$.

Specifically, the additional matching tasks for big cooperative learning is 
\beq\label{eq:model_to_model_all}
%\resizebox{0.99\columnwidth}{!}{$
	\min_{\thetav} \max_{\phiv} 
	\Ebb_{q(\Sbb^{1},\Tbb^{1})q(\Sbb^{2},\Tbb^{2})}
	\bigg[
	\bali
	& \Ebb_{p_{\thetav}\!(\xv_{\Tbb^{1}} | \xv_{\Sbb^{1}}) q(\xv_{\Sbb^{1}})} {\log} \sigma[f_{\phiv}\!(\xv;\Sbb^{2},\Tbb^{2}) \!-\! f_{\phiv}\!(\xv;\Sbb^{1},\Tbb^{1})] 
	\\
	& \!+\! \Ebb_{p_{\thetav}\!(\xv_{\Tbb^{2}} | \xv_{\Sbb^{2}}) q(\xv_{\Sbb^{2}})} {\log} \sigma[f_{\phiv}\!(\xv;\Sbb^{1},\Tbb^{1}) - f_{\phiv}\!(\xv;\Sbb^{2},\Tbb^{2})]
	\eali \bigg],
	%$}
\eeq
where the ``communication'' discriminator can be implicitly constructed with the same neural network $f_{\phiv}(\xv;\Sbb,\Tbb)$ from Eq. \eqref{eq:model_to_data_all}, as proved below.

Naively, one would develop the objective
\beq\label{appeq:model_to_model}
	\min_{\thetav} \max_{\phiv} 
	\left\{\bali 
		& \Ebb_{p_{\thetav}(\xv_{\Tbb^{1}} | \xv_{\Sbb^{1}}) q(\xv_{\Sbb^{1}})} {\log} \sigma[f'_{\phiv}(\xv;\Sbb^{1},\Tbb^{1},\Sbb^{2},\Tbb^{2})]
		\\
		& + \Ebb_{p_{\thetav}(\xv_{\Tbb^{2}} | \xv_{\Sbb^{2}}) q(\xv_{\Sbb^{2}})} {\log} (1-\sigma[f'_{\phiv}(\xv;\Sbb^{1},\Tbb^{1},\Sbb^{2},\Tbb^{2})]),
	\eali\right.
\eeq
where the optimal $f'_{\phi^{*}}(\xv;\Sbb^{1},\Tbb^{1},\Sbb^{2},\Tbb^{2}) = \log \frac{p_{\thetav}(\xv_{\Tbb^{1}} | \xv_{\Sbb^{1}}) q(\xv_{\Sbb^{1}})}{p_{\thetav}(\xv_{\Tbb^{2}} | \xv_{\Sbb^{2}}) q(\xv_{\Sbb^{2}})}$ is quite complicated for modeling.
Instead, one can alternatively leverage the $f_{\phiv}(\xv;\Sbb,\Tbb)$ from Eq. \ref{eq:model_to_data_all} (or Eq. \ref{eq:model_to_data_all_app}) to implicitly construct it, because
$$\bali
	f'_{\phi^{*}}(\xv;\Sbb^{1},\Tbb^{1},\Sbb^{2},\Tbb^{2})
	& = \log \frac{q(\xv_{\Sbb^{2}\cup\Tbb^{2}})}{p_{\thetav}(\xv_{\Tbb^{2}} | \xv_{\Sbb^{2}}) q(\xv_{\Sbb^{2}})}
	- \log \frac{q(\xv_{\Sbb^{1}\cup\Tbb^{1}})}{p_{\thetav}(\xv_{\Tbb^{1}} | \xv_{\Sbb^{1}}) q(\xv_{\Sbb^{1}})}
	\\
	& = f_{\phi^{*}}(\xv;\Sbb^{2},\Tbb^{2}) - f_{\phi^{*}}(\xv;\Sbb^{1},\Tbb^{1})
\eali$$

Accordingly, Eq. \eqref{appeq:model_to_model} is reformulated as 
\beq\label{appeq:model_to_model_v1}
\min_{\thetav} \max_{\phiv} 
\left\{\bali 
	& \Ebb_{p_{\thetav}(\xv_{\Tbb^{1}} | \xv_{\Sbb^{1}}) q(\xv_{\Sbb^{1}})} {\log} \sigma[f_{\phiv}(\xv;\Sbb^{2},\Tbb^{2}) - f_{\phiv}(\xv;\Sbb^{1},\Tbb^{1})] 
	\\
	& + \Ebb_{p_{\thetav}(\xv_{\Tbb^{2}} | \xv_{\Sbb^{2}}) q(\xv_{\Sbb^{2}})} {\log} \sigma[f_{\phiv}(\xv;\Sbb^{1},\Tbb^{1}) - f_{\phiv}(\xv;\Sbb^{2},\Tbb^{2})], 
\eali\right.\eeq
which is Eq. \ref{eq:model_to_model_all}. Accordingly, we conclude the proof.

\subsection{Training Settings of BigLearn-GAN}
\label{appsec:BigLearn_GAN_training}

\begin{table}[H]
	\centering
	\caption{Hyperparameters employed in the BigLearn-GAN experiments.}
	%	\resizebox{0.95\columnwidth}{!}{
		\begin{tabular}{l c c}
			\hline \hline
			Dataset & MNIST & CelebA  \\
			\hline \hline
			Image size & 64 & 120  \\
			Patch size & 8 & 10  \\
			\hline
			G-Encoder depth & 6 & 6  \\
			G-Encoder \#heads & 8 & 8  \\
			G-Encoder dim & 256 & 256  \\
			\hline 
			G-Decoder depth & 6 & 6  \\
			G-Decoder \#heads & 8 & 8  \\
			G-Decoder dim & 512 & 512  \\
			\hline 
			D depth & 6 & 6  \\
			D \#heads & 8 & 8  \\
			D dim & 256 & 256  \\
			\hline 
			GP \cite{mescheder2018training} & real & real  \\
			$\lambda_{\text{GP}}$ & 10 & 50  \\
			Learning rate & $10^{-4}$ & $10^{-4}$  \\
			Batch size & 256 & 128  \\
			%			Gradient clip & 5.0 & 5.0  \\
			\hline 
			Source ratio $\nicefrac{\|\Sbb\|}{\|\Lbb\|}$ & Beta(0.5,3) & Beta(0.5,3)  \\
			Target ratio $\nicefrac{\|\Tbb\|}{\|\Lbb\backslash\Sbb\|}$ & Beta(3,0.5) & Beta(3,0.5)  \\
%			Communication source ratio $\nicefrac{\|\Sbb^2\|}{\|\Sbb^1\cup\Tbb^1\|}$ & Beta(0.5,3) & Beta(0.5,3)  \\
			\hline \hline
		\end{tabular}
		%	}
	\label{apptab:hyperparam_settings}
\end{table}

%MNIST [-1,1]
%
%%input_size 24
%%patch_size 4 
%
%GNoise  GIt    GLastNorm LR
%
%%embed_dimG 128
%%depthG 3 
%%num_headsG 4
%%decoder_embed_dimG 256
%%decoder_depthG 3 
%%decoder_num_headsG 4 
%mlp_ratioG 2 
%
%DST  DIn  DLastNorm LN
%
%%embed_dimD 128
%%depthD 3 
%%num_headsD 4 
%mlp_ratioD 2 
%
%--D_no_padST 
%
%%setGP real
%%lambdaGP 1
%%lr 1e-4
%%batch_size 64 
%
%%Sratio -1 
%%Tratio -1 
%%CommuSratio -1 
%
%%SampleSratio  beta (0.5, 3)
%%SampleTratio  beta (3, 0.5)
%%SampleCommuSratio  beta (0.5, 3)

% ----------------------------------

%CelebA [-1,1]
%
%%input_size 120
%%patch_size 15 
%
%GNoise  GIt    GLastNorm LR
%
%%embed_dimG 256
%%depthG 6
%%num_headsG 8
%%decoder_embed_dimG 512
%%decoder_depthG 6
%%decoder_num_headsG 8
%mlp_ratioG 2 
%
%DST  DIn  DLastNorm LN
%
%%embed_dimD 256
%%depthD 6 
%%num_headsD 8
%mlp_ratioD 2 
%
%--D_no_padST 
%--toImg vit
%
%%setGP real
%%lambdaGP 10
%%lr 1e-4
%%batch_size 128 
%
%%Sratio -1 
%%Tratio -1 
%%CommuSratio -1 
%
%%SampleSratio  beta (0.5, 3)
%%SampleTratio  beta (3, 0.5)
%%SampleCommuSratio  beta (0.5, 3)

%\begin{table}%[H]
%	\centering
%	\caption{Input settings for Mnist and CelebA.}
%	\resizebox{0.95\columnwidth}{!}{
	%		\begin{tabular}{l c c c c c c c c c}
		%			\hline \hline
		%			Dataset & image/size & patch/size & depth/Gencoder & depth/Gdecoder & depth/D & embeddim/Gencoder & embeddim/Gdecoder & embeddim/D & num/heads
		%			\\ \hline 
		%			Mnist & 32 & 4 & 3 & 3 & 3 & 128 & 256 & 128 & 4
		%			\\
		%			CelebA & 120 & 10 & 6 & 6 & 6 & 256 & 512 & 256 & 8
		%			\\ \hline \hline
		%		\end{tabular}
	%	}
%	\label{apptab:input_settings}
%\end{table}

We employ the same network architectures in the previous section for both experiments on the MNIST and CelebA datasets, with the detailed hyperparameters summarized in Table \ref{apptab:hyperparam_settings}.
The AdamW optimizer \cite{loshchilov2017decoupled} with $\beta=(0.1, 0.999)$ is employed, with constant learning rates used for updating both the generator and the discriminator.
We follow Algorithm \ref{alg:BigLearn-GAN} for big cooperative learning.
The experiments are conducted with a NVIDIA GeForce RTX $3090$ GPU with a $24$GB memory.
An MNIST experiment takes half a day, while a CelebA one takes about $2-3$ days.
Code will be released upon publication.

\begin{algorithm}[H]
	\caption{Big Cooperative Learning Generative Adversarial Nets (BigLearn-GAN)}
	\label{alg:BigLearn-GAN}
	\begin{algorithmic}[1]
		\Require Training data $\xv\sim q(\xv)$, maximum number $N_{\text{o}}$ of outer iterations, maximum number $N_{\text{in}}$ of inner iterations.
		%		\vspace{1mm}
		\Ensure Consistent and stable local optima, the generator $\thetav^{*}$ and discriminator $\phiv^{*}$.
		
		%		\vspace{1mm}
		\State Randomly initialize $\phiv$ and $\thetav$
		
		\While{iter $\le N_{\text{o}}$}
		%		\vspace{1mm}
		\State Sample a $(\Sbb,\Tbb)$ pair from $q(\Sbb,\Tbb)$
		
		\State \# Update the discriminator parameters $\phiv$
		\State \qquad Calculate the loss $J(\phiv)$ based on Eq. \ref{eq:model_to_data_all}
		\State \qquad Update $\phiv \leftarrow \phiv - \nabla_{\phiv} J(\phiv)$
		\\\Comment{\texttt{often regularized by the gradient penalty \cite{mescheder2018training}}}
		
		\State \# Update the generator parameters $\thetav$
		\State \qquad Calculate the loss $L(\thetav)$ based on Eq. \ref{eq:model_to_data_all}
		\State \qquad Update $\thetav \leftarrow \thetav - \nabla_{\thetav} L(\thetav)$

		\EndWhile
	\end{algorithmic}
\end{algorithm}

Overall, we find that it's straightforward to implement the BigLearn-GAN on the MNIST dataset, without resorting to additional training ``tricks.''
However, on the more complicated CelebA dataset, we find it challenging to stabilize the BigLearn-GAN training with a Transformer-based generator and discriminator, similar to what's shown in the ViTGAN \cite{lee2021vitgan}.
To stabilize the training, we additionally
$(i)$ strengthen the gradient penalty (GP) \cite{mescheder2018training} with a GP coefficient $\lambda_{\textbf{GP}}=50$;	
$(ii)$ adopt L$2$ attention \cite{kim2021lipschitz} to improve Lipschitzness of the ViT discriminator;
$(iii)$ employ $3$ additional convolutional layers in the head of the generator;
$(iv)$ concatenate the same set of noise embeddings layer-wisely into the G-Decoder layers;
and
$(v)$ use a larger hyperparameter $\epsilon=10^{-5}$ in the AdamW optimizer.
Although we manage to achieve realistic generation on CelebA (as shown in Sections \ref{sec:VersatileCapabilities_biglearning} and \ref{appsec:add_results_BigLearn-GAN}), we feel that there are still unsolved mysteries in the stable training of the BigLearn-GAN when using a Transformer-based generator and discriminator.

%--DUseLips --DSimDemodProj --ProbBLData 0.5 --glr 1e-4 --dlr 1e-4 --weight_decay 0.01 --clip_gradG -1 --clip_gradD -1 --toImg 3convin --max_iter 500000 --lambdaGP 50 --Ninner 5

\subsection{Additional Experimental Results of BigLearn-GAN}
\label{appsec:add_results_BigLearn-GAN}

More experimental results, complementing the limited demonstrations of the main manuscript, are given below. Please refer to the captions for details.

\begin{figure}[H]
	\centering
	\includegraphics[width=1.\columnwidth]{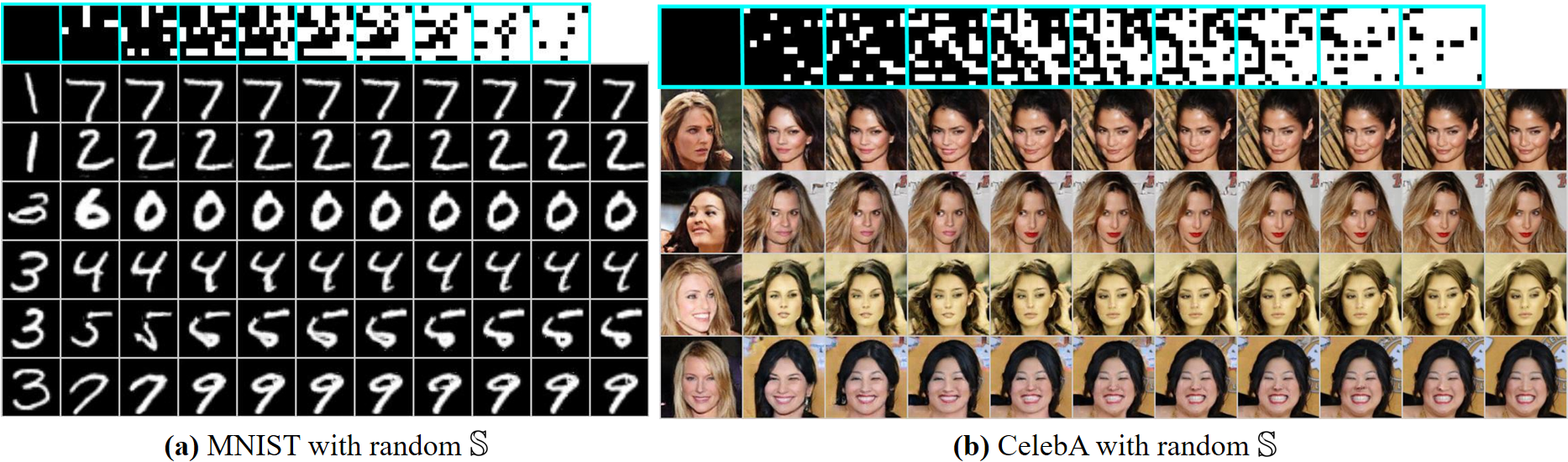}
	\includegraphics[width=1.\columnwidth]{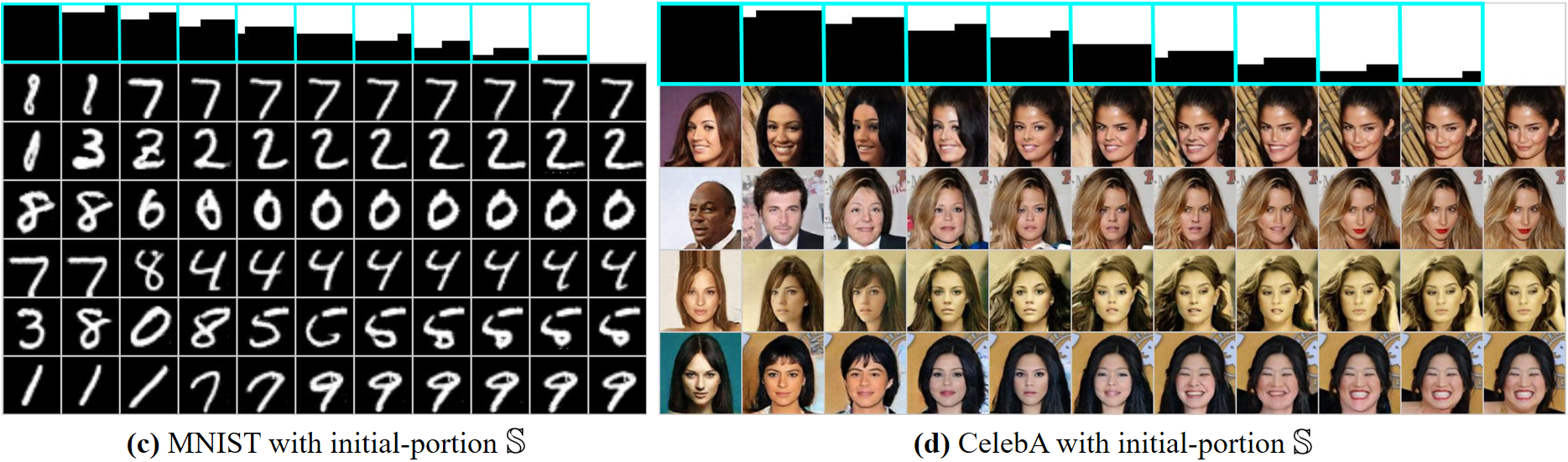}
	\vspace{-3mm}
	\caption{Demonstrating the generation/completion capabilities of big learning when gradually increasing the ratio of $\Sbb$ from $0$ (joint generation) to $0.9$, from left to right.
		Shown in the light-blue boxes of the first row are the masks of $\xv_{\Sbb}$ applied in each column; white/black indicates $\Sbb/\Tbb$. 
		The right-most column shows ground-truth $\xv$ shared in each row.
		Note each row also employs the same noise.
		It's clear that the generations become increasingly similar/dissimilar to the ground-truth $\xv$ as the ratio of $\Sbb$ increases/decreases, as expected. 
		See the category, style, and thickness of the MNIST generations as the ratio of $\Sbb$ decreases, as well as the identity, expression, hairstyle, and gender of the CelebA generations.
		%		With fewer source patches, most of the information are controlled by noise and the generation shows different identity, hairstyle expression (CelebA), class, shape from $\xv$. 
		%		The generation are diverse in identity, hairstyle, expression, shape of jaw. One can observe that the generation shows rich variation in hairstyle, expresion, identity, and gender.
		%		A similar diverse generation is also observed on MNIST, \eg the class, shape, thickness.
		Big learning produces realistic and diverse generations/completions in all situations.
	}
	\label{fig:diff_Sratio}
\end{figure}

\begin{figure}[H]
	\centering
	\includegraphics[width=1\columnwidth]{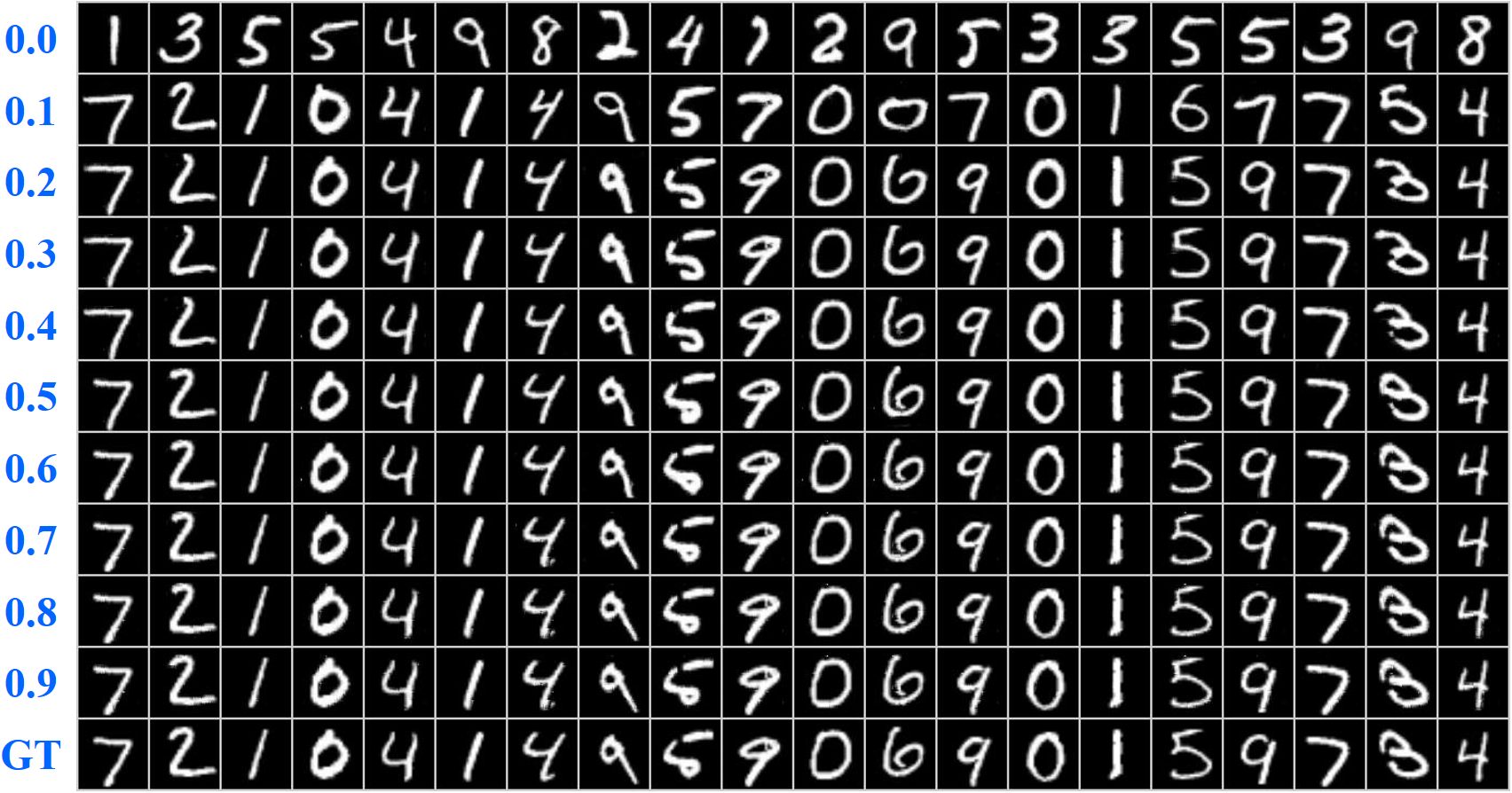}
	\caption{More MNIST generations/completions from big learning when gradually increasing the ratio of $\Sbb$ from $0.0$ to $0.9$.
	}
	\label{fig:}
\end{figure}

\begin{figure}[H]
	\centering
	\includegraphics[width=1\columnwidth]{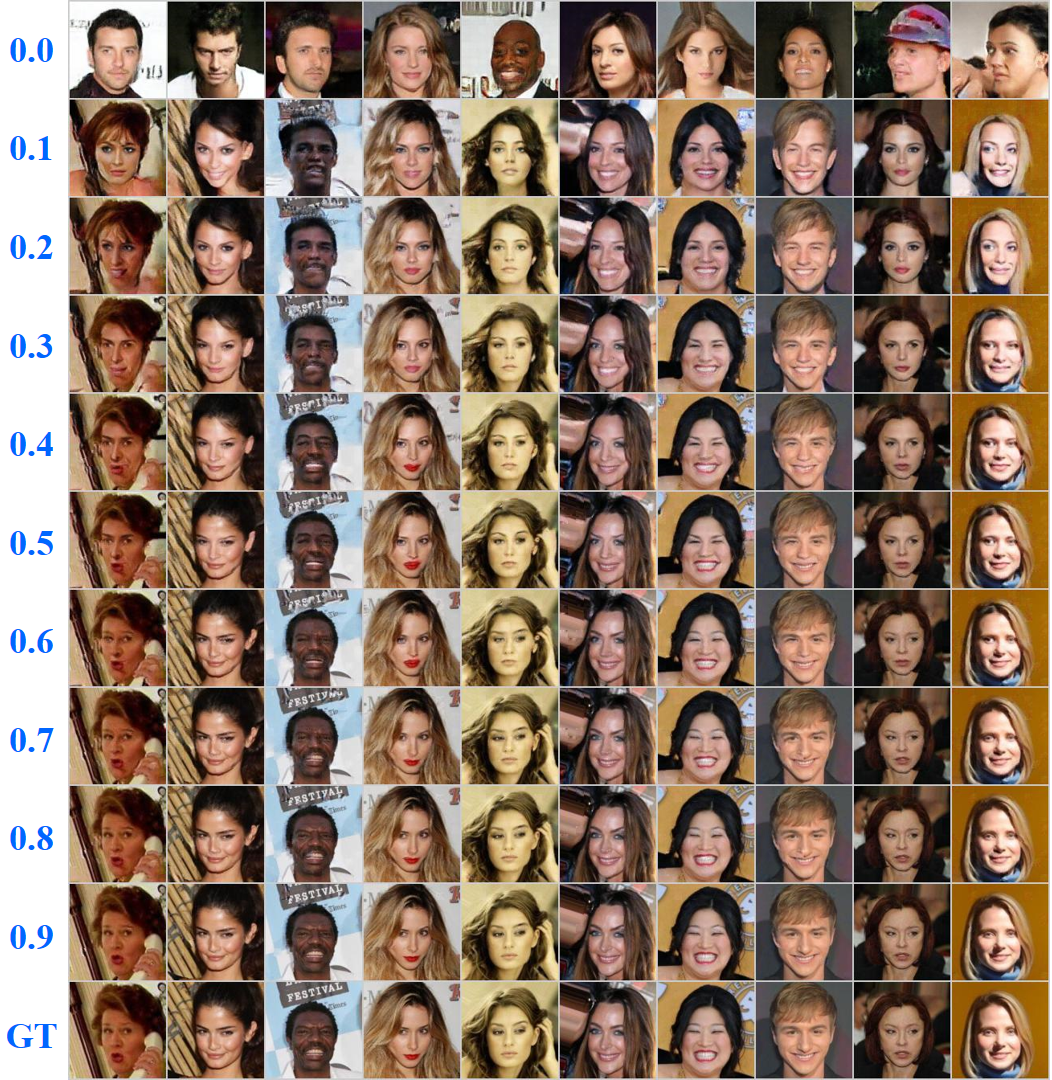}
	\caption{More CelebA generations/completions from big learning when gradually increasing the ratio of $\Sbb$ from $0.0$ to $0.9$.
	}
	\label{fig:}
\end{figure}

\begin{figure}[H]
	\centering
	\includegraphics[width=1.2\columnwidth]{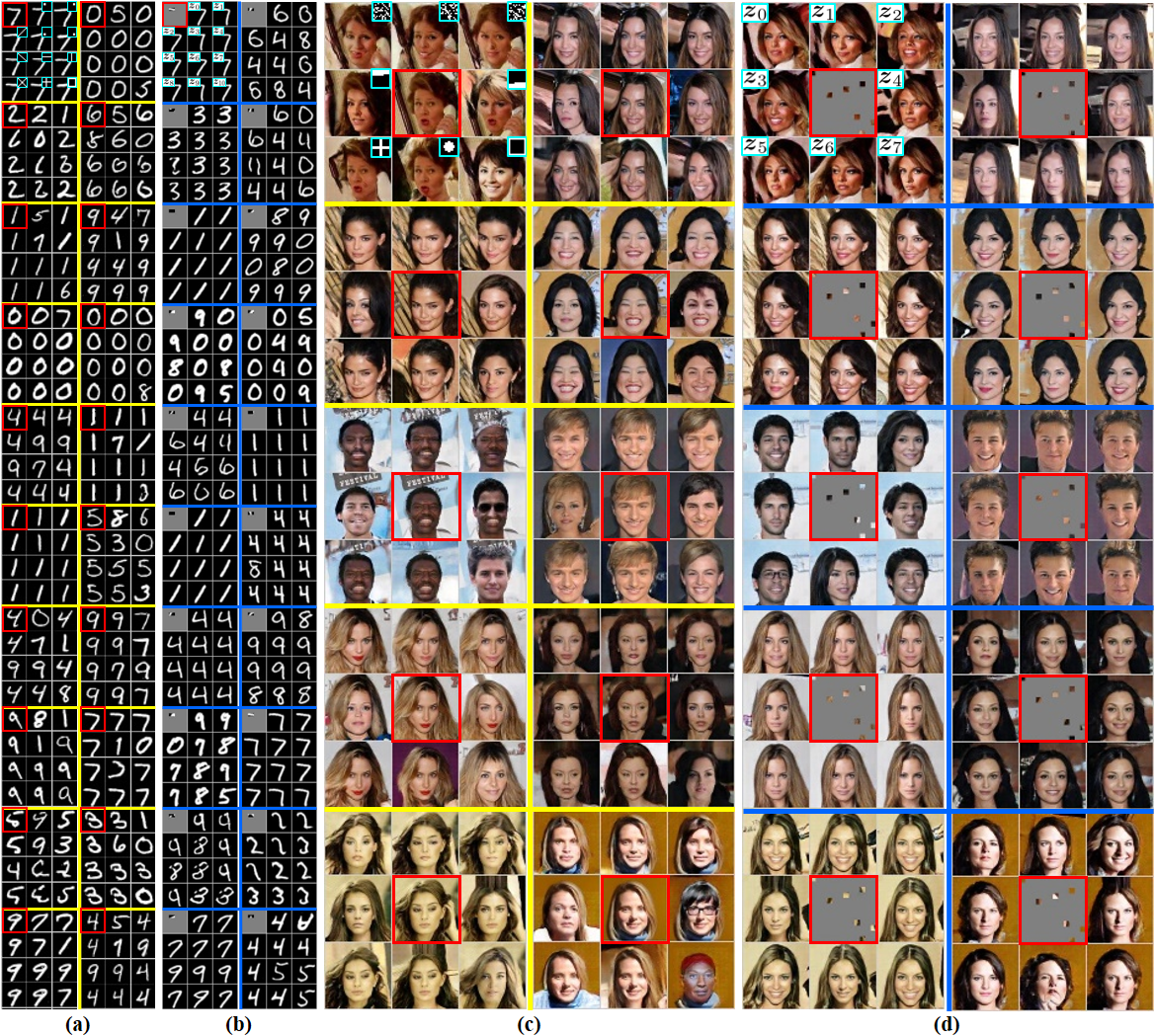}
	\caption{The diverse generations/completions of big learning with (a)(c) various $\Sbb$ settings and (b)(d) different noises. 
		Shown in red boxes are either the ground-truth images $\xv$ or the source $\xv_{\Sbb}$. 
		Big learning delivers diverse realistic generations \wrt different $\Sbb$/noise settings.
	}
	\label{fig:}
\end{figure}

\section{Big Cooperative Learning in Multi-Modal Applications}
\label{appsec:biglearn_MultiModal}

The following experiments are conducted with a NVIDIA GeForce RTX $3090$ GPU with a $24$GB memory. The total running time is about several days.

\subsection{Leveraging Big Learning to Unify Classification and Generation}
\label{appsec:biglearn_genclass}

We follow \cite{bao2021beit,ramesh2021zero} to vector-quantize an image into discrete tokens $\xv' \in \Zbb^{(L-1)\times 1}$, which is then combined with the label $y' \in \{1, \cdots, C\}^{1\times 1}$ to constitute a multi-modal data $\xv\in \Zbb^{L\times 1}$. 
We employ no transformations (\ie $\Xv=\xv$) and construct the universal Transformer-based $p_{\thetav}(\xv_{\Tbb}|\xv_{\Sbb})$ with the two-stream attention \cite{yang2019xlnet} to model the diverse dependencies of $\xv_{\Tbb}$ on $\xv_{\Sbb}$ for various $(\Sbb,\Tbb)$s.
Finally, we implement big cooperative learning of Definition \ref{def:biglearn} in the maximum-likelihood-learning territory via
\beq\label{eq:bl_xlnet}
	\max_{\thetav} \Ebb_{q(\zv)q(t)q(\xv_{\zv_{<t}})} \Ebb_{q(x_{z_{t}}|\xv_{\zv_{<t}})} \log p_{\thetav}(x_{z_t}|\xv_{\zv_{<t}}),
\eeq
where $q(\zv)$ is a user-defined distribution of the random permutation $\zv$, $\Sbb=\{\zv_{<t}\}$, $\Tbb=\{z_t\}$, and often $p_{\thetav}(x_{z_t}|\xv_{\zv_{<t}}) =\text{Categorical}(x_{z_t}|\Pv_{\thetav}(\xv_{\zv_{<t}}))$ is modeled as a categorical distribution with probabilities $\Pv_{\thetav}(\xv_{\zv_{<t}})$.
With specific settings, the objective in Eq. \eqref{eq:bl_xlnet} may recover the objective in \cite{yang2019xlnet}.

\begin{figure}[H]
	%	\vspace{-6mm}
	\centering
	\subfloat[Joint Generation]{
		\includegraphics[height=0.39\columnwidth]{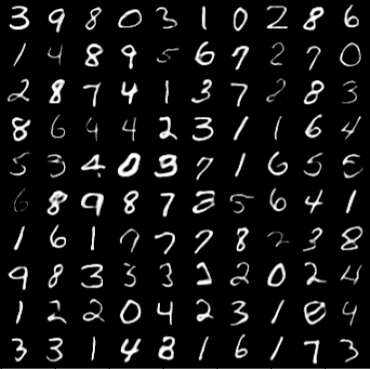}
		\label{fig:}}
	\qquad\quad
	\subfloat[Label-Conditioned Generation]{
		\includegraphics[height=0.39\columnwidth]{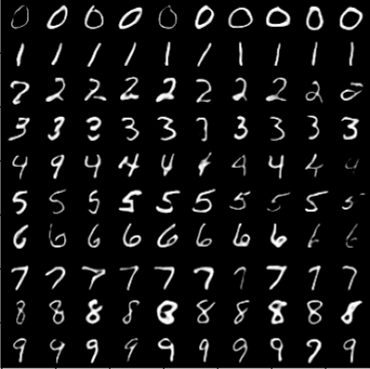}
		\label{fig:}}
	\caption{Demonstration of versatile data-sampling capabilities of big cooperative learning in multi-modal scenarios.
		Capabilities are retrieved from the universal $p_{\thetav}(\xv_{\Tbb}|\xv_{\Sbb})$ with specified $(\Sbb, \Tbb)$.
		With $\Sbb=\{1,\cdots,L-1\}$ and $\Tbb={L}$, the capability $p_{\thetav}(y'|\xv')$ delivers a testing accuracy of $95.27\%$.
	}
	\label{appfig:biglearning_genclass_all}
	%	\vspace{-2mm}
\end{figure}

As demonstrated in Fig. \ref{appfig:biglearning_genclass_all}, it's evident that, after big cooperative learning, the universal $p_{\thetav}(\xv_{\Tbb}|\xv_{\Sbb})$ simultaneously delivers versatile capabilities like joint generation $p_{\thetav}(\xv')$, label-conditioned generation $p_{\thetav}(\xv'|y')$, classification $p_{\thetav}(y'|\xv')$, \etc 
proving that big cooperative learning is capable of delivering versatile \emph{cross-modal} capabilities with one universal model.
Note these simultaneously-delivered capabilities are likely valuable for counterfactual analysis and reasoning.

\subsection{Using Big Cooperative Learning as a New Fine-Tuning Strategy}
\label{appsec:biglearn_finetune}

Instead of naively leveraging big cooperative learning to pretrain a foundation model from scratch (which is extremely expensive), we consider using it as a special fine-tuning strategy, to demonstrate its effectiveness in a lightweight manner.

Specifically, we design experiments based on the Hugging Face transformers library \cite{wolf-etal-2020-transformers}, the GLUE benchmark \cite{wang2018glue}, and the XLNET \cite{yang2019xlnet} that outperforms the BERT on many NLP tasks.
We employ the same pretrained \texttt{xlnet-base-cased} model and continually train it on the downstream RTE/MRPC/SST-2 classification tasks via ($i$) the naive fine-tuning (\ie identical to the original XLNET \cite{yang2019xlnet}, termed FT) and ($ii$) big cooperative learning (termed big-learn), respectively. 
As a multi-modal data $\xv\in \Zbb^{L\times 1}$ here also consists of tokens $\xv' \in \Zbb^{(L-1)\times 1}$ and a label $y' \in \{1, \cdots, C\}^{1\times 1}$, other settings are set similar to Section \ref{appsec:biglearn_genclass}.

The objectives for both FT and big-learn are detailed below. 
\begin{itemize}
	\item \textbf{FT.} 
	For fine-tuning on classification applications, often a cross-entropy loss is employed, which is identical to 
	\beq
	\Lc_{\text{FT}}(\thetav) = \Ebb_{q_{\text{downstream}}(\xv,y)} [-\log p_{\thetav}(y|\xv)],
	\eeq
	where $q_{\text{downstream}}(\xv,y)$ is the distribution of the data from the downstream classification task.
	
	\item \textbf{Big-learn.}
	For direct comparisons, we formalize the big-learn objective as 
	\beq\label{appeq:big_learn_1}
		\Lc_{\text{big-learn}}(\thetav) = \Lc_{\text{FT}}(\thetav) + \beta_{\text{BigLearn}} \Lc(\thetav),
	\eeq
	where $\beta_{\text{BigLearn}}$ is a hyperparameter and $\Lc(\thetav)$ is the loss in Eq. \eqref{eq:bl_xlnet}. 
	For other hyperparameters, we simply reuse those from Section \ref{appsec:biglearn_genclass}, without careful tuning.
\end{itemize}

\begin{table}[H]
	\centering
	\caption{Tested hyperparameters when comparing FT with big-learn on the GLUE benchmark.}
	\resizebox{\columnwidth}{!}{
		\begin{tabular}{l c c c c}
			\hline \hline
			{Task}$\backslash${Hyperparameter} & Learning Rate & \#Epochs & WarmUp Steps & $\beta_{\text{BigLearn}}$
			\\ \hline 
			RTE & [2e-5, 4e-5, 6e-5] & [3, 4, 7, 10, 15] & [0, 120] & [0., 0.2, 0.4, 0.6, 0.8] 
			\\
			MRPC & [2e-5, 4e-5, 6e-5] & [3, 4, 7, 10, 15] & [0, 120] & [0., 0.2, 0.4, 0.6, 0.8] 
			\\
			SST-2 & [2e-5, 4e-5, 6e-5] & [2, 3, 4] & [0, 1200] & [0., 0.2, 0.4] 
			\\ \hline \hline
		\end{tabular}
	}
	\label{apptab:glue_hyperpara}
\end{table}

We extensively compare FT with big-learn on the downstream RTE/MRPC/SST-2 classification tasks, by evaluating the accuracy and/or F1 score on the Dev set across the combinations of the tested hyperparameters shown in Table \ref{apptab:glue_hyperpara}. 
These hyperparameters are chosen following \cite{devlin2018bert,yang2019xlnet}.

\begin{table}[htb]
	\centering
	\caption{Big learning serves as a superior fine-tuning strategy.
		The best/median metrics are calculated among the combinations of the tested hyperparameters of Table \ref{apptab:glue_hyperpara}.}
	%	\vspace{2mm}
	\resizebox{0.6\columnwidth}{!}{
		\begin{tabular}{l | c c | c c}
			\hline \hline
			\multirow{2}{*}{Task} & \multicolumn{2}{c|}{Best Accuracy / F1} & \multicolumn{2}{c}{Median Accuracy / IQR}
			\\ 
			& FT & big-learn & FT & big-learn
			\\ \hline 
			RTE & 71.84  & $\textbf{75.09}$ & 66.06/2.34  & $\textbf{70.75/1.44}$ 
			\\
			MRPC & 88.97/92.09 & $\textbf{90.20/93.03}$ & 87.00/2.45 & $\textbf{87.74/1.10}$  
			\\
			SST-2 & 94.15  & $\textbf{95.18}$ & 93.75/0.45 & $\textbf{94.66/0.28}$  
			\\ \hline \hline
		\end{tabular}
	}
	\label{apptab:glue_ACC}
	%	\vspace{-2mm}
\end{table}

\begin{figure}[H]
%	\vspace{-5mm}
	\centering
	\subfloat[RTE]{
		\includegraphics[height=0.33\columnwidth]{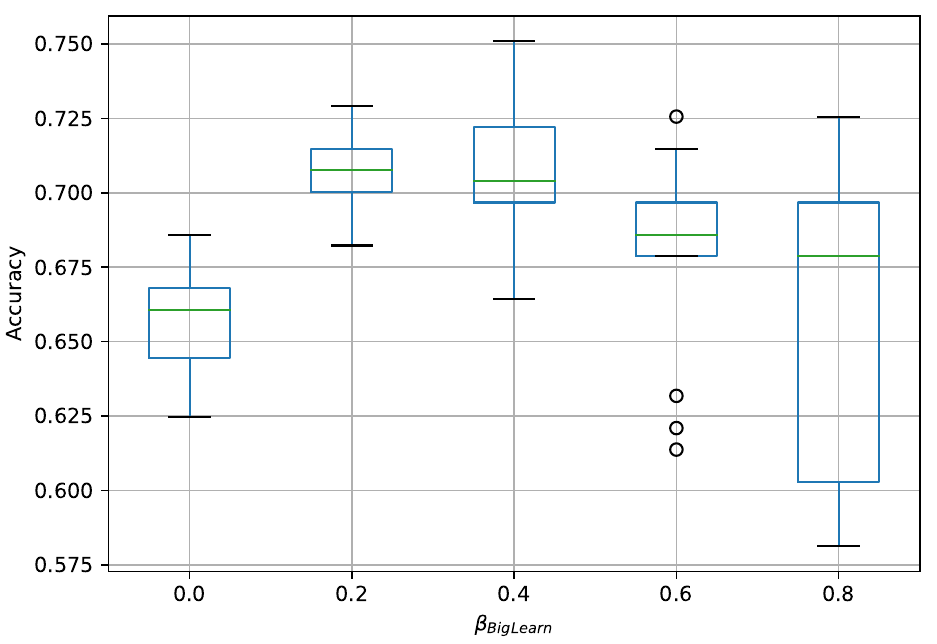}
		\label{fig:}}
	\subfloat[MRPC]{
		\includegraphics[height=0.33\columnwidth]{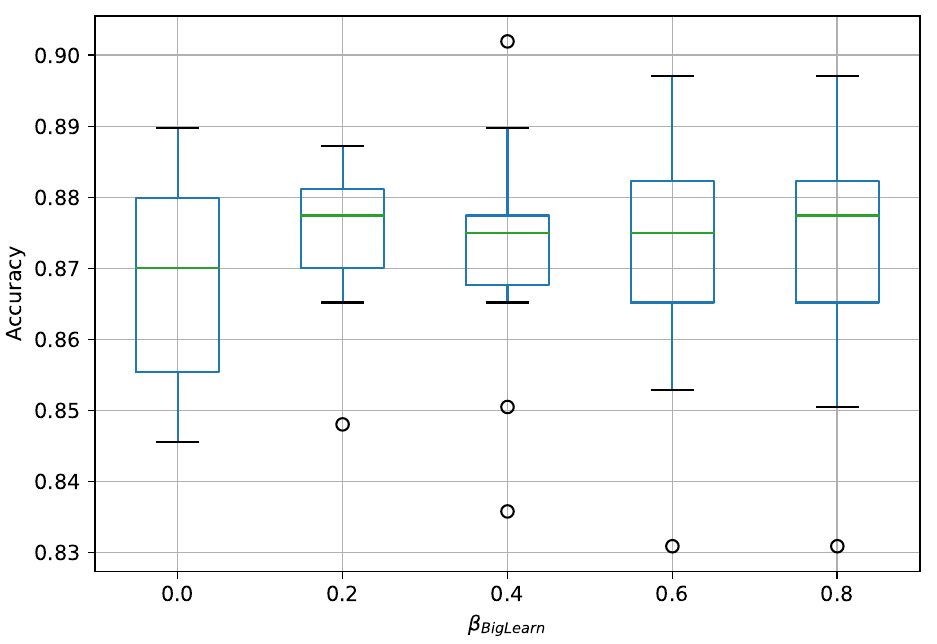}
		\label{fig:}}
	\qquad
	\subfloat[SST-2]{
	\includegraphics[height=0.33\columnwidth]{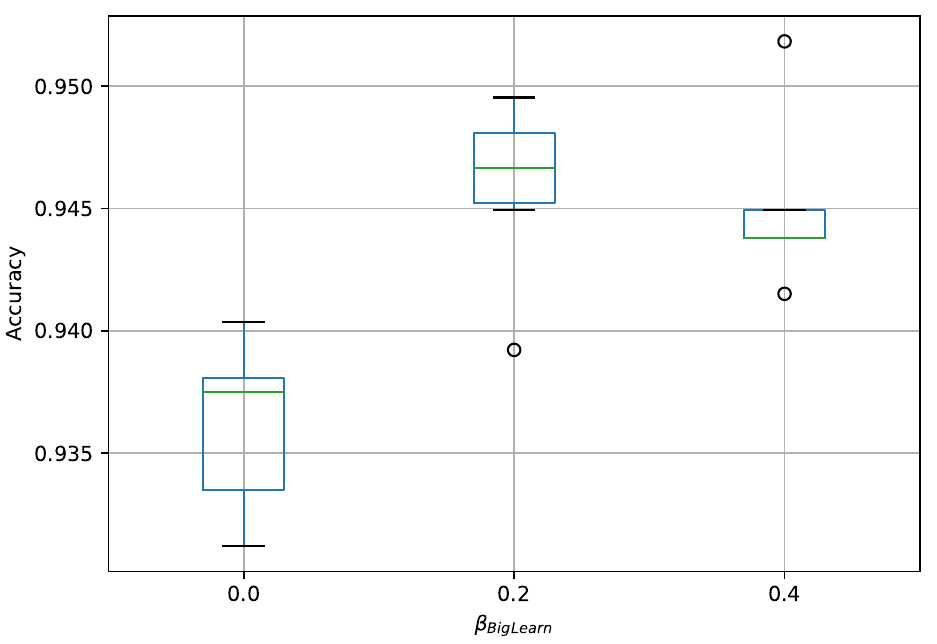}
	\label{fig:}}
	\caption{Boxplots of the Dev-set accuracies from FT and our big-learn. Note big-learn with $\beta_{\text{BigLearn}}=0$ is identical to FT (see \eqref{appeq:big_learn_1}).
		It's clear that big-learn consistently outperforms FT on all three tasks.
	}
	\label{appfig:boxplot_glue}
%	\vspace{-2mm}
\end{figure}

The best/median metrics are summarized in Table \ref{apptab:glue_ACC} and Fig. \ref{appfig:boxplot_glue} shows the corresponding boxplots; it's clear that our big-learn consistently outperforms FT. Accordingly, the big learning can serve as a superior fine-tuning strategy than the naive FT.
It's worth highlighting we did not carefully tune our big-learn; therefore, it's likely that its performance could be further improved.

We'd like to emphasize that the big learning can reduce the pretrain-finetuning gap because 
\begin{itemize}
	\item it can serves as the objective for both pretraining and finetuning;
	\item one can even rely on the big learning to completely merge the pretraining and finetuning phases, leading to a zero gap from the perspective of learning.
\end{itemize}

\section{Limitations}
\label{appsec:limitations_big_learn}

The main assumptions of big cooperative learning are listed at the beginning of Section \ref{sec:our_method} of the main manuscript.
However, those assumptions might be violated in practice, potentially resulting in unexpected behavior of the presented big cooperative learning.
\begin{itemize}
	\item We assume trustworthy data samples from $q(\xv)$. Although this assumption is widely employed, it's often violated in practice, \eg due to limited accessibility to \iid samples, unintentional interventions in data collection and preprocessing, or even hostile attack. Careful data engineering should be able to significantly alleviate the violation.
	
	\item We assume sufficient model capacity of $p_{\thetav}(\Xv_{\Tbb}|\Xv_{\Sbb}), \forall (\Sbb,\Tbb)$. This is another common issue of deep-learning methods. Accordingly, we cannot justify this assumption in the experiments of Sections \ref{sec:VersatileCapabilities_biglearning} and \ref{sec:MultiModal_BigLearn_exp}. 
	The simulations of Sections \ref{sec:2D_gmm_simulation} and \ref{sec:25GMM_exploration} should ease the concern.
	
\end{itemize}

In addition to the above common limitations, the presented big cooperative learning, as a general concept widely applicable in diverse domains, are not comprehensively verified in practical applications in this paper. 
For example, we cannot verify its effectiveness in training a large language model like the GPT4 \cite{GPT4} because it would be too computationally expensive.
We focus on presenting the big-cooperative-learning concept and explicitly demonstrating its principle via tailored controllable simulations (and preliminary experiments on median-size practical applications) in this paper, and leave as future work a comprehensive verification of its effectiveness in practical applications.

In the BigLearn-GAN experiment on the CelebA dataset, although we manage to deliver versatile realistic generations (as shown in Sections \ref{sec:VersatileCapabilities_biglearning} and \ref{appsec:add_results_BigLearn-GAN}), we believe that there are still unsolved mysteries in terms of an appropriate modeling of the generator/discriminator of the BigLearn-GAN and its stable adversarial training.
These issues are not uncommon in the GAN community and we will endeavour to address them in future research.

Big cooperative learning, for now, is data-driven.
Although it's possible to leverage its flexibilities (in constructing versatile matchings across diverse transformed domains with complete and/or incomplete data) to introduce certain domain knowledge (\eg data augmentations during preprocessing), we haven't made the corresponding contributions in this paper. 
We leave that as future research.

\end{document}